\documentclass[10pt,twocolumn,letterpaper]{article}

\usepackage{cvpr}              
\usepackage[normalem]{ulem}
\usepackage[table,xcdraw]{xcolor}
\usepackage{comment}
\usepackage{fontawesome}

\usepackage{multirow}
\usepackage{pifont}

\definecolor{myred}{RGB}{196,64,60}
\definecolor{mygreen}{RGB}{0,155,85}
\definecolor{myblue}{RGB}{46,134,193}

\definecolor{mygray}{gray}{0.6}
\definecolor{mygray-bg}{gray}{0.9}

\definecolor{cvprblue}{rgb}{0.21,0.49,0.74}
\usepackage[pagebackref,breaklinks,colorlinks,allcolors=cvprblue]{hyperref}

\newcommand{\modelname}{SpatialLLM}
\newcommand{\datasetname}{SpatialVQA}
\definecolor{ablatecolor}{gray}{.60}
\definecolor{lightyellow}{RGB}{255, 255, 200}
\definecolor{lightorange}{RGB}{255,200,150}

\def\paperID{14934} 
\def\confName{CVPR}
\def\confYear{2025}

\title{\modelname: A Compound 3D-Informed Design towards\\Spatially-Intelligent Large Multimodal Models}

\makeatletter
\def\thanks#1{\protected@xdef\@thanks{\@thanks
        \protect\footnotetext{#1}}}
\makeatother

\author{\textbf{Wufei Ma}$^1$, \textbf{Luoxin Ye}$^1$, 
\textbf{Nessa McWeeney}$^1$,
\textbf{Celso M de Melo}$^2$,
\textbf{Jieneng Chen}$^{1,}$\textsuperscript{\faEnvelopeO}\thanks{\textsuperscript{\faEnvelopeO} Corresponding author: Jieneng Chen (\href{mailto:jchen293@jh.edu}{\textsc{jchen293@jh.edu}})},
\textbf{Alan Yuille}$^1$ \\
\small $^1$Johns Hopkins University \quad $^2$DEVCOM Army Research Laboratory
}

\begin{document}
\maketitle
\begin{abstract}
Humans naturally understand 3D spatial relationships, enabling complex reasoning like predicting collisions of vehicles from different directions. Current large multimodal models (LMMs), however, lack of this capability of 3D spatial reasoning. This limitation stems from the scarcity of 3D training data and the bias in current model designs toward 2D data. 
In this paper, we systematically study the impact of 3D-informed data, architecture, and training setups, introducing \modelname, a large multi-modal model with advanced 3D spatial reasoning abilities. 
To address data limitations,  we develop two types of 3D-informed training datasets: (1) 3D-informed probing data focused on object's 3D location and orientation, and (2) 3D-informed conversation data for complex spatial relationships. Notably, we are the first to curate VQA data that incorporate 3D orientation relationships on real images.
Furthermore, we systematically integrate these two types of training data with the architectural and training designs of LMMs, providing a roadmap for optimal design aimed at achieving superior 3D reasoning capabilities.  Our \modelname\xspace advances machines toward highly capable 3D-informed reasoning, surpassing GPT-4o performance by 8.7\%. Our systematic empirical design and the resulting findings offer valuable insights for future research in this direction. Our project page is available \href{https://3d-spatial-reasoning.github.io/spatial-llm/}{here}.
\end{abstract}
    
\section{Introduction}
\label{sec:intro}

When humans observe a scene, they perceive more than isolated objects; they intuitively understand the spatial relationships among these objects~\cite{spelke2007core}, enabling complex reasoning and interaction with the environment. For example, recognizing that a car is approaching an obstacle and might collide involves interpreting the 3D object orientations and spatial relationships. This innate ability to comprehend and reason about 3D space is crucial for navigating and interacting with the physical world.

\begin{figure}[!ht]
  \centering
  \includegraphics[width=\columnwidth]{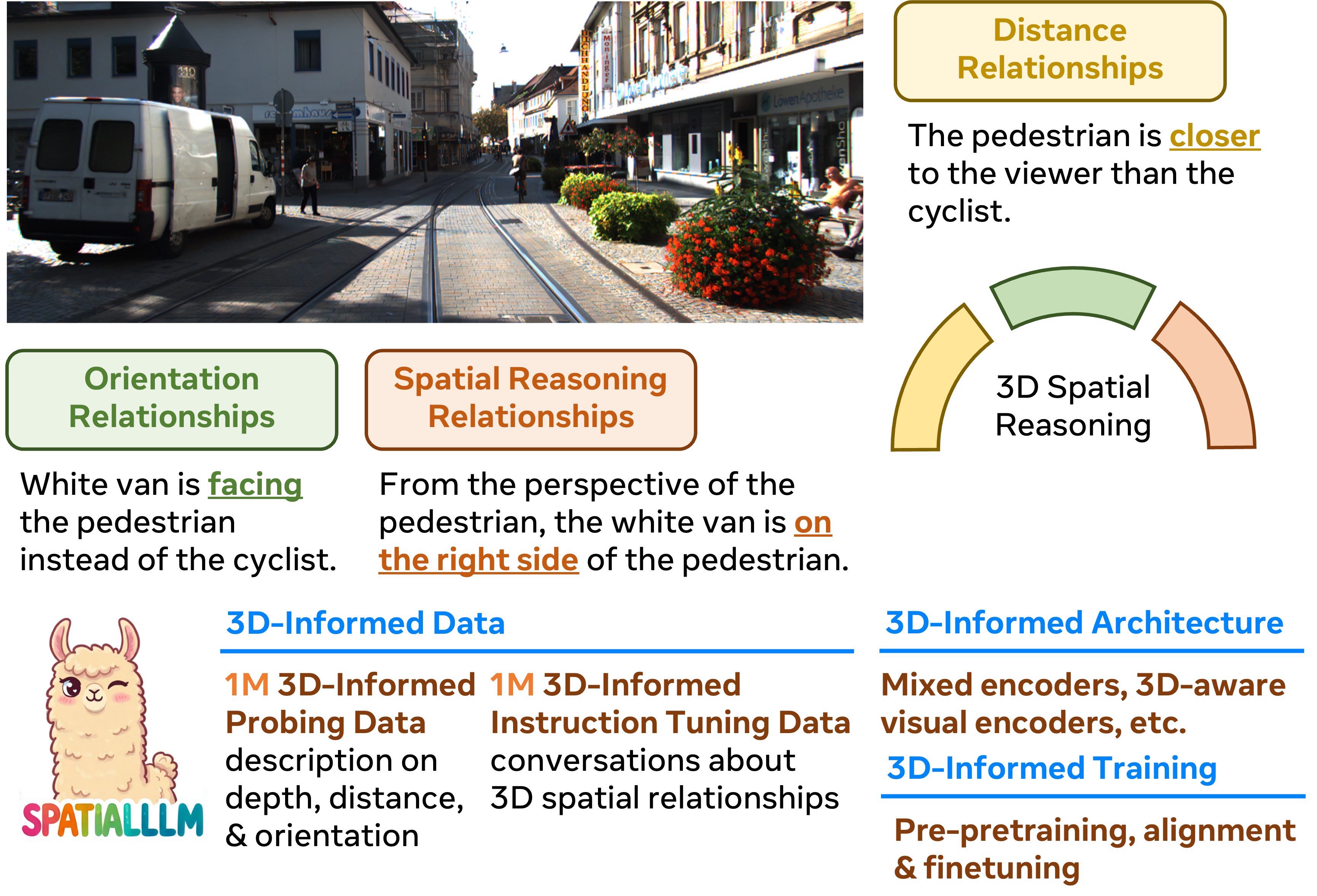}
  \caption{
  3D spatial reasoning is crucial for LMMs to ground objects in 3D space and infer their 3D spatial relationships, such as distance, orientation, and spatial interactions. We systematically study the impact of 3D-informed data, architecture, and training setups, introducing \modelname, an LMM with advanced 3D spatial reasoning abilities.
  }
  \label{fig:teaser}
\end{figure}

Despite significant advancements, current  LMMs~\cite{achiam2023gpt, team2023gemini, claude, liu2024improved, liu2024visual}, exhibit limited capability of 3D spatial reasoning. These models excel at identifying objects and generating descriptive captions but struggle with tasks that require intricate 3D reasoning, such as predicting object interactions or understanding physical events.

The primary challenge lies in the scarcity of high-quality 3D spatial relationship data, which prevents models from learning effective 3D spatial reasoning.
Collecting a small set of high-quality 3D-aware data to tackle the first challenge is feasible, albeit labor-intensive, using readily available tools. 
Recent efforts~\cite{chen2024spatialvlm, spacellava} leverage  vision foundation models to recognize~\cite{zhang2024recognize}, detect~\cite{liu2023grounding}, segment~\cite{kirillov2023segment} and estimate the depth~\cite{bhat2023zoedepth,yang2024depth} in images.  These approaches form semi-automated pipelines to construct pseudo-2.5D datasets for training LMMs with spatial reasoning capabilities, such as depth ordering~\cite{tong2024cambrian, chen2024spatialvlm}.

However, a significant gap remains: previous works~\cite{chen2024spatialvlm, spacellava, cheng2024spatialrgpt} have primarily focused on 3D distance relationships, overlooking the crucial role of 3D object orientation. This is primarily due to the absence of robust 6D pose estimators. Understanding object orientation is essential for complex spatial reasoning tasks, such as determining if two cars are facing the same direction to avoid dangerous collision. 
Addressing this gap, we aim to incorporate 3D orientation relationships—converted from ImageNet3D~\cite{deng2009imagenet}—into our data engine, making us the first to enable complex spatial reasoning involving 3D orientation relationships.

Even with 3D-informed data in hand, fully unlocking its potential remains challenging due to the intricate, multi-stage development process required for an LMM. For example, SpatialVLM~\cite{chen2024spatialvlm} utilizes 3D-informed data during the instruction tuning stage, but it remains uncertain whether such data can enhance other stages like vision encoder pretraining or multimodal alignment. This limitation suggests that a more holistic approach is necessary.

In this paper, we present a systematic investigation into building LMMs with enhanced 3D spatial reasoning capabilities through a compound design approach, as illustrated in Fig.~\ref{fig:teaser}. Our key contributions are threefold:

First, we identify three fundamental types of 3D spatial relationships that LMMs must master: distance relationships, orientation relationships, and complex spatial reasoning that combines both. To evaluate these capabilities, we introduce \datasetname{}, a comprehensive benchmark containing 1,323 questions that target 3D spatial understanding, built upon diverse urban and indoor scenes from Omni3D~\cite{brazil2023omni3d}.

Second, we propose a novel compound 3D-informed design that introduces improvements across multiple dimensions, leading to our proposed \textbf{\modelname{}} model. On the data front, we develop two types of 3D-informed training datasets: (1) 3D-informed probing data focused on object-level 3D properties, and (2) 3D-informed conversation data for for complex spatial relationships. Architecturally, we explore mixing different visual encoders and investigate their 3D awareness capabilities. For training, we introduce 3D-informed data at multiple stages: pre-pretraining, multimodal alignment, and instruction tuning. 

Third, we present the first comprehensive search over the LMM design space for spatial reasoning tasks and propose a roadmap towards developing state-of-the-art models in this domain. The experiment demonstrates that our proposed \textbf{\modelname{}} achieves state-of-the-art performance on spatial reasoning tasks, surpassing both GPT-4o~\cite{achiam2023gpt} (by 8.7\%) and spatial-centric models like SpatialVLM~\cite{chen2024spatialvlm} (by 10.5\%). Our systematic analysis reveals that while architectural improvements contribute a 1.3\% gain, the strategic incorporation of 3D-informed data across different training stages yields a substantial 13.7\% performance improvement.

Our findings highlight the importance of a holistic approach to improving LMMs' 3D spatial reasoning capabilities, demonstrating that both architectural choices and carefully curated open-source 3D-informed training data play crucial roles in achieving superior performance. This work establishes a new benchmark for spatial reasoning in LMMs and provides valuable insights for future research in this direction.
\section{Related Works}
\label{sec:related}

\textbf{Large multimodal models (LMMs)}. Recently, there has been a significant surge in research on multimodal LLMs, exemplified by products like GPT-4~\cite{achiam2023gpt} and Gemini~\cite{team2023gemini}, as well as open-source models such as LLaVA~\cite{liu2024visual,liu2024improved, li2024llava} and Qwen-VL~\cite{bai2023qwenvl}. 
These open-source models combine pre-trained vision encoders~\cite{radford2021learning,jia2021scaling,oquab2023dinov2,chen2024vitamin} with LLMs~\cite{touvron2023llama2, bai2023qwen,jiang2023mistral}, enabling the language models to handle a wide range of tasks that involve images as input. 
Typically, these methods has the alignment stage that aligns image features with the input embedding space of the language model using connectors~\cite{liu2024improved, cha2024honeybee,li2022blip}, trained with paired images and captions~\cite{sharma2018conceptual}. 
The instruction fine-tuning~\cite{liu2024visual} stage uses data that contains both images and instructions to enhances LMMs. 
However, current LMMs are unsuccessful in spatial visual reasoning, due to limited spatial-aware data and stagnant 3D-informed architecture design. We aim to bridge this research gap.

\textbf{3D spatial reasoning with LMMs}.
Understanding spatial relationships between objects in images is crucial for visual reasoning. 
While this comes naturally to humans, LMMs were found to struggle with recognizing spatial relations~\cite{rajabi2024gsr,wang20243d,wang2024compositional}, due to a lack of 3D-aware spatial-related data during training. While neural symbolic methods~\cite{wang20243d} achieved improved performance on synthetic benchmarks by involving explicit 3D reasoning modules~\cite{ma2022robust,jesslen2024novum}, they struggled to generalize to a broader range of objects found in real images.
Fortunately, with foundation models like SAM~\cite{kirillov2023segment} and depth estimators~\cite{bhat2023zoedepth,yang2024depth}, semi-automatic annotation for 3D-aware VQA data has been adopted in recent studies~\cite{chen2024spatialvlm,cheng2024spatialrgpt}. SpatialVLM~\cite{chen2024spatialvlm} uses the pseudo-annotated VQA data for training only while SpatialRGPT~\cite{chen2024spatialvlm,cheng2024spatialrgpt} further augments LLaVA with extra input of regional mask and depth map during inference, which is not our focus.
Although human-annotated 3D-aware VQA data is rare, we aim to leverage human-annotated 3D object-level datasets (like ImageNet3D~\cite{ma2024imagenet3d}) by simply converting image-pose pairs into image-text pair data.
However, even with the collected data, a major challenge remains in effectively utilizing 3D-aware data to train optimal models. Current approaches like SpatialVLM~\cite{chen2024spatialvlm} use the standard LLaVA and two-stage training.
To date, no work has addressed when (which training stage) and where (which network component) are optimal to inject the 3D-awareness to LMMs.
We bridge this gap by conducting a compound design that simultaneously considers 3D-aware data, architecture, and the training, leading to a best-performing design.
Rather than competing, our study has the potential to complement prior works~\cite{chen2024spatialvlm,cheng2024spatialrgpt} by offering new compound design recipe.
\section{Methods} \label{sec:methods}

In this section we start by reviewing LMMs in \cref{sec:methods-preliminary}, specifically the architecture and the choice of training data and strategies. We present the task of reasoning 3D spatial relationships and explain the challenges LMMs face when answering these questions in \cref{sec:methods-3DSR}.
Then we study how to address these challenging problems with LMMs and introduce our \modelname{} in \cref{sec:methods-3di}.
%

\subsection{Preliminary of LMMs} \label{sec:methods-preliminary}

\textbf{Data}. The success of deep learning in vision is largely due to well-curated data~\cite{sun2017revisiting,gadre2023datacomp}, and LMMs similarly depend on this foundation, albeit with diverse data formats below.
\begin{itemize}[leftmargin=*,itemsep=0pt]
\item Image-only data, such as ImageNet~\cite{deng2009imagenet}, is commonly used for MAE~\cite{he2022masked}/DINOv2~\cite{oquab2023dinov2} pretraining with a self-supervised objective. This step enables the model to learn rich visual features solely from visual signals.
\item Noisy image-text pairs: Large-scale image-text pairs~\cite{schuhmann2022laion,gadre2023datacomp,li2024if} are used for pretraining with contrastive loss (e.g., CLIP~\cite{radford2021learning, yu2022coca}) to align visual and textual representations. 

\item Multimodal alignment: At this stage, image-caption pairs (e.g., CC3M~\cite{sharma2018conceptual}) are utilized to align visual features with the language model's word embedding space, typically through image-captioning tasks. This pretraining ensures that the model can interpret and relate visual content to textual queries in a shared semantic space. 

\item Multimodal instruction-following data. Visual instruction tuning data is essential but challenging to collect, as it rarely exists in its natural form online. Prior works~\cite{liu2024visual, liu2024improved, mckinzie2024mm1, tong2024cambrian} repurpose VQA benchmarks~\cite{krishna2017visual, goyal2017making} into instruction-tuning datasets.

\end{itemize}
\textbf{Architecture}. A standard LMM~\cite{liu2024visual, liu2024improved} consists of a visual encoder to process the image, a multimodal connector to transform the visual feature to visual token, and a LLM for reasoning. However, the extent to which these architectural components contribute to the spatial understanding of LMMs remains unknown. We review these architectural components as below.
\begin{itemize}[leftmargin=*,itemsep=0pt]
\item Vision encoder: Most LMMs rely on language-supervised models like CLIP~\cite{radford2021learning}, leveraging web-scale noisy image-text data. Recent LMMs~\cite{lu2024deepseek, tong2024cambrian}  integrate vision models trained only on visual signals, including self-supervised~\cite{zhou2021ibot, oquab2023dinov2, he2022masked} and generative models~\cite{rombach2022high}.
\item Connector: Features from a visual encoder are mapped into the LLM token space by connectors~\cite{alayrac2022flamingo, li2022blip, cha2024honeybee, liu2024visual} such as a MLP~\cite{liu2024visual, liu2024improved}. 
\item LLMs: LLMs~\cite{achiam2023gpt,touvron2023llama2, bai2023qwen}  serve as the foundational backbone of LMMs, driving their reasoning capabilities. Open-source LLMs like Llama series~\cite{touvron2023llama2, dubey2024llama} are frequently adopted to enhance the reasoning capability.
\end{itemize}
\textbf{Training}. The aforementioned data supports pre-pretraining for strong visual representation learning, alignment pretraining to synchronize visual and language representations, and instruction tuning to enable multimodal instruction-following and reasoning capabilities
\begin{itemize}[leftmargin=*,itemsep=0pt]
\item Pre-pretraining. This stage focuses on developing foundational visual representations, often with reconstruction-based objectives (e.g., MAE~\cite{he2022masked}, DINOv2~\cite{oquab2023dinov2}). This step prepares the model to handle more complex multimodal tasks by first refining its visual understanding.
\item Multimodal alignment pretraining. At this stage, the model is trained to describe images in details to align visual and language representations in the same space. 
\item Visual instruction tuning. The final stage involves refining the model's ability to process complex multimodal instructions using visual instruction tuning data. This often incorporates transformed VQA datasets, enhancing the model’s capacity in handling tasks that demand coordinated reasoning across language and vision.
\end{itemize}

\subsection{Reasoning 3D Spatial Relationships} \label{sec:methods-3DSR}

3D spatial reasoning analyzes the spatial relationships between objects in the 3D world space, such as objects being far or close to each other, or one object lies to the left or right of another object~\cite{wang20243d}. It extends beyond standard 3D tasks, such as 3D object detection and 3D pose estimation that studies the 3D nature of individual objects, and focuses on understanding their mutual relationships in 3D. In this work we identify three main types of 3D spatial relationships, \ie, distance, orientation, and spatial reasoning, and study how we can better address these problems with LMMs.

\paragraph{3D spatial relationships.} For \textit{distance spatial relationships}, we study if models can estimate and compare 3D distances between objects. These spatial relationships require more than depth awareness, which gives the distance between the viewer and the object and can be roughly estimated from the object scale, but also the 3D distances between two objects that are often harder to estimate.
On the other hand, \textit{orientation spatial relationships} require models to understand the 3D orientations of an object, \eg, predicting if two objects are facing the same direction.
Lastly, \textit{spatial reasoning spatial relationships} look into 3D spatial relationships that need a combination of 3D awareness of location and orientation and then performing spatial reasoning on the 3D information.

\begin{figure*}[t]
    \centering
    \includegraphics[width=\textwidth]{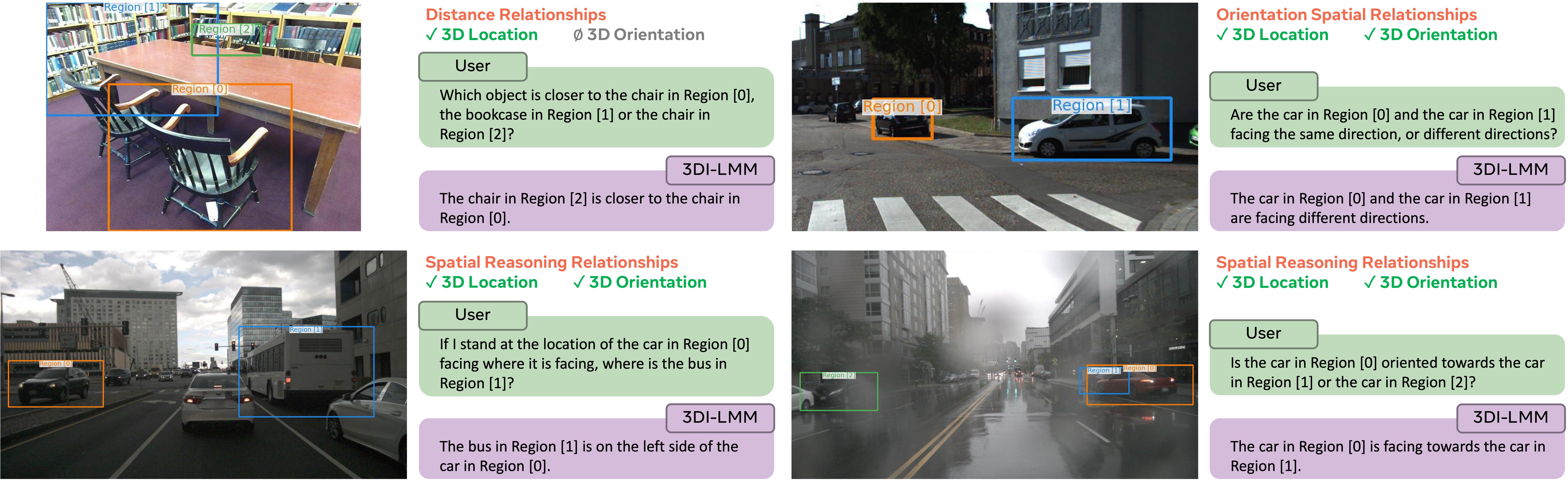}
    \caption{Examples from our \datasetname{} benchmark featuring a broad range of questions that require 3D spatial reasoning.}
    \label{fig:spatialvqa}
\end{figure*}

\subsubsection{Challenges of 3D spatial reasoning} \label{sec:methods-3DSR-challenges}
Reasoning about 3D spatial relationships remains a challenging task for state-of-the-art open-sourced and proprietary LMMs. We attribute this to two main limitations in modern LMMs, the \textit{lack of 3D awareness} and \textit{inability to perform 3D spatial reasoning}.

\paragraph{3D awareness} refers to the ability of a visual encoder to represent 3D-aware features such as 3D object shapes and 3D orientations. Previous studies resorted to proxy tasks, \eg, part correspondence~\cite{el2024probing} and pose estimation~\cite{ma2024imagenet3d}, to quantitatively evaluate the 3D awareness of visual foundation models with linear probing. Results showed that most visual encoder, specifically ones trained with language supervision, demonstrated limited 3D awareness of 3D object poses~\cite{ma2024imagenet3d}, which inevitably limits the 3D spatial reasoning capabilities of LMMs built upon these encoders.
           
\paragraph{Inability to perform 3D spatial reasoning.} To infer the 3D spatial relationships, models must also be able to perform 3D spatial reasoning from 3D-aware visual features. Despite the abundant Internet-scale data for multi-modal training, we argue that high-quality 3D spatial reasoning data are scarce. Existing pretraining and visual instruction tuning data for LMMs~\cite{liu2024visual,tong2024cambrian} focused on detailed descriptions and conversations about scenes, appearances, and actions, while being vague about the 3D spatial relationships that build up the high-level physical events. Despite their success on a wide range of image understanding tasks, these LMMs fails to learn the nature of 3D scenes or to perform 3D spatial reasoning.

\subsubsection{\datasetname \xspace for  Evaluation}
\label{sec:build_spatialvqa}

Despite the rising interests in spatial reasoning, there lacks a comprehensive open source benchmark for reasoning 3D spatial relationships. Early datasets on spatial reasoning were built on 3D scans rather than images~\cite{azuma2022scanqa,ye20213d}. Recent visual-language benchmarks on spatial reasoning either focused on 2D spatial relationships~\cite{cheng2024spatialrgpt,chen2024spatialvlm}, \eg, left or right in the image plane, or only on distance spatial relationships~\cite{tong2024cambrian}.

To enable a thorough evaluation of the 3D spatial reasoning capabilities of LMMs, we introduce \datasetname, a visual question answering benchmark that covers a wide variety of 3D spatial relationships. Our \datasetname{} distinguishes itself from all previous spatial reasoning benchmarks in the sense that all questions require different levels of 3D awareness and cannot be answered from 2D spatial reasoning only.
Specifically, \datasetname{} consists of 1,323 questions, ranging from questions can be answered from distance-awareness only, \eg, which object is closer to a third object, to questions that require estimating objects' 3D orientations, \eg, if two objects are facing the same direction, and eventually questions that require complex 3D spatial reasoning over a combination of object 3D locations and orientations, \eg, an object is in the front/left/back/right direction of another object.

We build our \datasetname{} on images from Omni3D~\cite{brazil2023omni3d}, with 3D bounding box annotations on diverse objects from both urban~\cite{caesar2020nuscenes,geiger2012we} and indoor scenes~\cite{song2015sun,baruch2021arkitscenes,hypersim}. We follow \cite{cheng2024spatialrgpt,tong2024cambrian} and develop rule-based methods to generate visual question-answer pairs from the 3D groundtruths. Please refer to \cref{fig:spatialvqa} for some examples of \datasetname{} and to appendix Sec.B for details about our benchmark.

\subsection{Compound 3D-Informed Design} \label{sec:methods-3di}

\paragraph{Problem formulation.} While previous approaches focused on generating 3D-informed instruction tuning data to enable LMMs with the abilities to estimate distances, depths, or spatial relationships~\cite{chen2024spatialvlm,cheng2024spatialrgpt}, it is largely understudied the best recipe for developing LMMs with strong 3D spatial reasoning capabilities. Motivated by the two aforementioned limitations of LMMs in \cref{sec:methods-3DSR-challenges}, we consider two main aspects in our compound 3D-informed design -- the architecture design that leads to visual encoders with strong 3D awareness and the training setup that advances the 3D spatial reasoning capabilities of LMMs.

\subsubsection{Design space}
\label{sec:design_space}
\begin{figure}[t]
  \centering
  \includegraphics[width=\columnwidth]{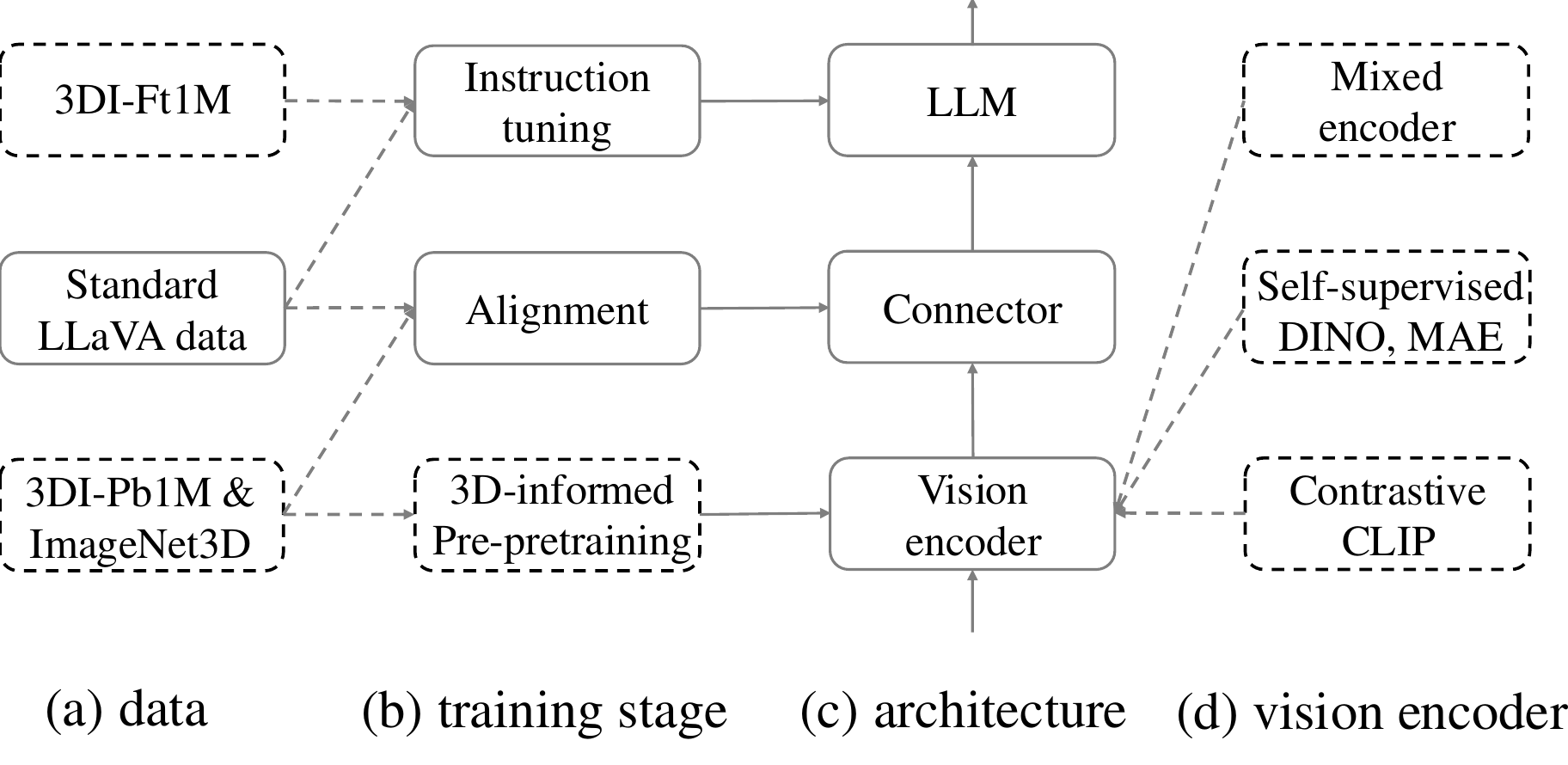}
  \caption{\textbf{Design space} for LMMs capable of spatial reasoning. \dashuline{The dashed boxes and lines} highlight our new design space compared to LLaVA-v1.5. This compound design simultaneously considers 3D-informed data, architecture, and training methods to search for the best-performing models for spatial reasoning.
  } 
\label{fig:design_space}
\end{figure}

We introduce the design space considered in our work, \ie, choices of training data, model architecture, and training setup that advance the 3D spatial reasoning capabilities of LMMs. Refer to \cref{fig:design_space} for an overview of our design space.

\begin{figure}[t]
  \centering
  \includegraphics[width=\columnwidth]{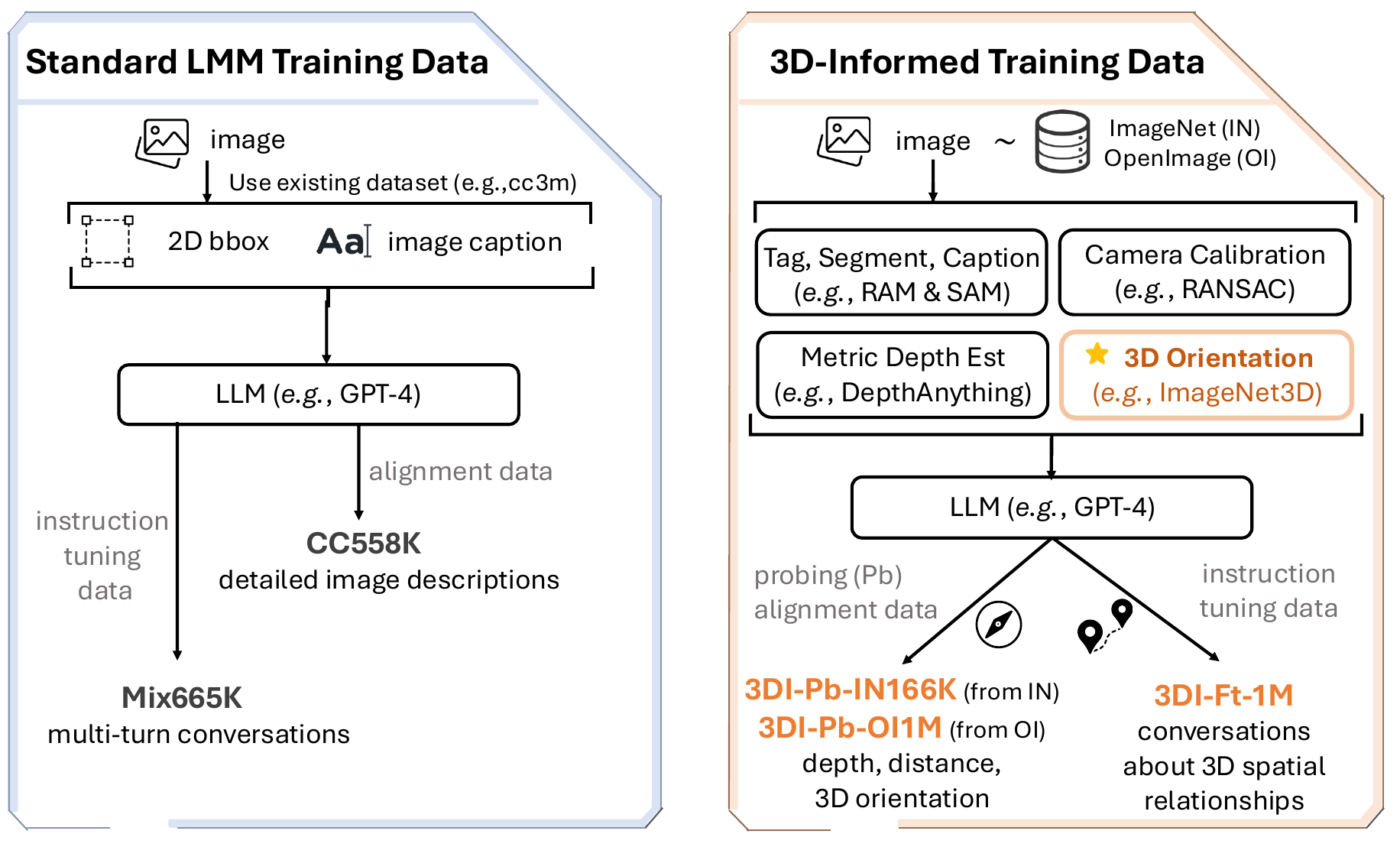}
   \caption{Comparison of data cards showcasing the curation process and data types: standard LLaVA data (left) vs. our 3D-informed data (right). For our 3D-informed data curation, we label depth and distance of objects in OpenImage~\cite{kuznetsova2020open} using semi-automatic tools, resulting in 3DI-Pb-OI1M. Beyond this, we introduce 3D orientation probing data derived from human-annotated ImageNet3D~\cite{ma2024imagenet3d}, resulting in 3DI-Pb-IN166K for alignment. We further augment the conversation about spatial relationship, leading to 3DI-Ft-1M for instruction tuning.
   }
   \label{fig:3di_data}
\end{figure}

\begin{figure*}[t]
  \centering
  \includegraphics[width=0.99\linewidth]{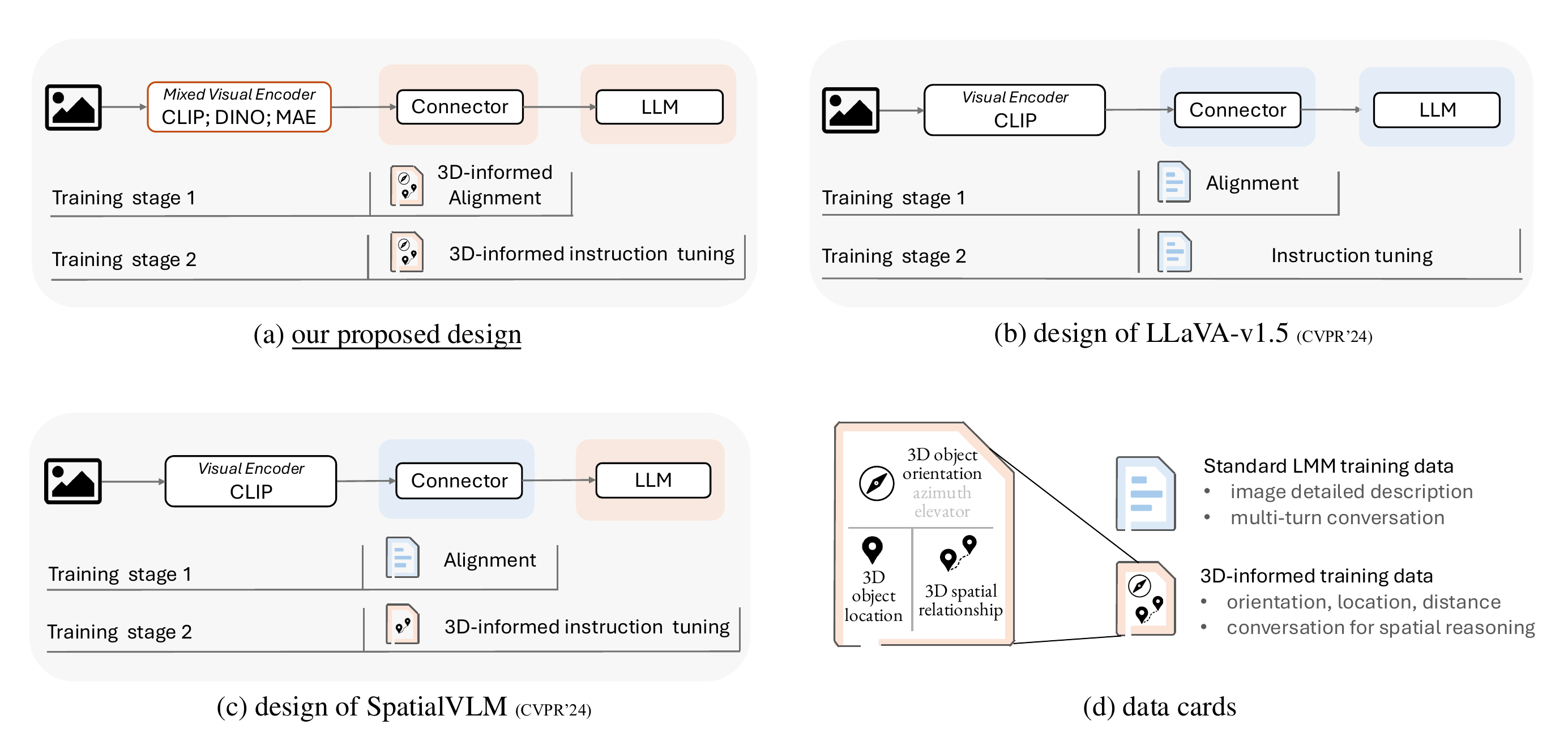}
  \caption{\textbf{Design instantiation and comparison}. (a)  Architecture and Training of our proposed design. We investigate the 3D-awareness of mixed visual encoders, and incorporate 3D-informed data at each training stage across all architecture components.   (b) LLaVA comparison, where only traditional data are used. (c)  SpatialVLM comparison, where 3D-informed data are used in the final stage. 
  (d) Traditional data vs. 3D-informed data. Note SpatialVLM lacks of 3D orientation data.
  Architectures trained with traditional data and 3D-informed data are marked with \textcolor{blue}{blue} and \textcolor{orange}{orange}  backgrounds, respectively.
  } 
  \label{fig:design_comparison}
\end{figure*}

\paragraph{Data.} Standard LMMs follow LLaVA~\cite{liu2024visual} and adopte feature alignment data for connector pretraining and visual instruction data for supervised instruction tuning. These datasets focus on detailed descriptions of images or machine-generated multi-turn conversations. Besides standard visual-language datasets, we further consider 3D-informed data, including 3D-aware probing data and 3D-informed instruction tuning data. Inspired by~\cite{ma2024imagenet3d}, the 3D-aware probing data involves \textit{fundamental} object-level 3D properties, such as the azimuth and elevation 3D viewpoint of an object, depth of the object from the viewer, and 3D distances between two objects. The goal of the 3D-aware probing data it to effectively inject fundamental 3D awareness into the models and benefit subsequent reasoning. Meanwhile, the 3D-informed instruction tuning data focuses on \textit{high-level} 3D spatial relationships between the objects, which are built on a combination of the fundamental 3D properties and 3D spatial reasoning.

To generate the 3D-aware probing data and the 3D-informed instruction tuning data, we follow the previous data generation pipelines~\cite{chen2024spatialvlm,cheng2024spatialrgpt} and exploit visual foundation models to extract and tag the objects in 3D space. Furthermore, we adopt a pretrained 6D pose estimator in \cite{ma2024imagenet3d} and estimate 3D orientations of the objects, which provides the crucial 3D information for generating 3D-informed data regarding various orientation-related 3D spatial relationships. We illustrate our 3D-informed data generation pipeline in \cref{fig:3di_data} and refer the readers to appendeix Sec.A for details of our 3D-informed data generation.

\paragraph{Architecture.} 
Inspired by Probe3D~\cite{el2024probing} which probes the 3D-awareness of visual foundation models, we study the design of visual encoder to enable better 3D spatial reasoning. 
We consider two types of visual encoders:
(i) Frozen \& pretrained visual encoder CLIP~\cite{radford2021learning} following~\cite{liu2024visual} but with the option to mix a wider variety of models, such as MAE~\cite{he2022masked} and DINOv2~\cite{oquab2023dinov2}, and 
(ii) 3D-aware visual encoders finetuned with our 3D-informed data, \eg, contrastive pretraining or finetuning with LoRA using language modeling loss.

\paragraph{Training setup.} To address the limitations of 3D-aware LMMs discussed in \cref{sec:methods-3DSR-challenges}, we propose new training setups that aim to improve 3D awareness and advance the 3D spatial reasoning capabilities. We consider the following: (i) in Stage 0: use 3D-informed probing data (3DI-Pb-OI1M/1N166K) to tune the Lora layers of CLIP vision encoder (Pre-pretraining); (ii) in Stage 1: 3D-informed multimodal alignment using a combination of standard CC558K~\cite{liu2024visual} and our 3D-informed probing data (3DI-Pb-OI1M/1N166K); and (iii) in Stage 2: 3D-informed instruction tuning with the standard Mix665K~\cite{liu2024visual} and our 3D-informed visual instruction tuning data (3DI-Ft-1M).


\subsubsection{Designing 3D-Informed LMM: A Roadmap}

%
\textbf{Design instantiation and comparison}.
The mentioned design space can result in several design instantiations, as depicted in Fig.~\ref{fig:design_comparison}.
In a multimodal LLM, a pre-trained CLIP visual encoder extracts grid features from the input image. These features are converted into visual tokens via a connector and then processed by the LLM for visual reasoning.
Fig.~\ref{fig:design_comparison} (b) shows that LLaVA is a standard design space without any new spatial-centric design, meaning that the 3D-awareness is completely missing in LLaVA (Fig.~\ref{fig:design_comparison} (b)). 
As in Fig.~\ref{fig:design_comparison} (c), SpatialVLM improves LLaVA by collecting 3D-informed data that is applied to instruction tuning for LLM, boosting the spatial reasoning capability. 
Compared to SpatialVLM, our design (in Fig.~\ref{fig:design_comparison} (a)) inject the 3D-awareness by introduce 3D-informed alignment with ImageNet3D data. Additionally, our 3D-informed intruction tuning data contain extra 3D orientation data that is missing in spatialVLM.

\begin{figure}[t]
  \centering
  \includegraphics[width=\columnwidth]{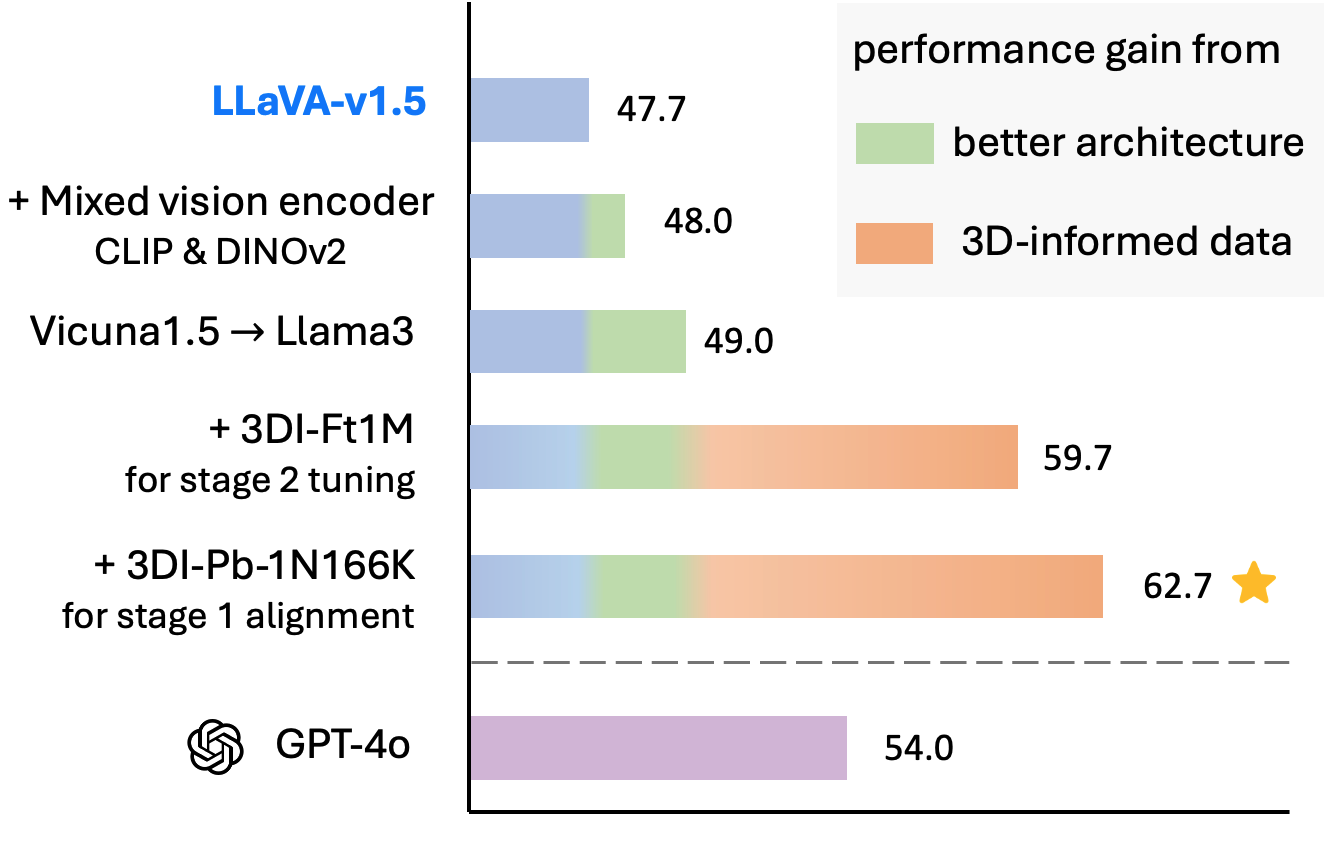}
  \caption{We modernize a standard LLaVA-v1.5 towards the design of a 3D-informed LMM. The bars show answer accuracies on SpatialVQA benchmark, indicating spatial reasoning capability of each model; results for the GPT-4o is shown with the purple bar.}
   \label{fig:search_roadmap}
\end{figure}

\textbf{Design roadmap}. We present the roadmap going from a standard LLaVA to our final model  in Fig.~\ref{fig:search_roadmap}. 
 We start with LLaVA-v1.5 which only achieves a 47.7\% accuracy on our \datasetname{}. This will be our baseline. We then study a series of design decisions.

\begin{itemize}[leftmargin=*,itemsep=0pt]
\item \textit{+ mixed vision encoder}. 
 To build stronger visual encoder, we mix the original CLIP vision encoder~\cite{radford2021learning} with the DINOv2~\cite{oquab2023dinov2} vision encoder to extract the hybrid feature maps, lifting a 0.3\% performance gain.  This show that the self-supervised feature can enhance spatial reasoning.
\item 
 \textit{Vicuna1.5$\rightarrow$Llama3}. We switch our large language model from Vicuna-v1.5-7B~\cite{zheng2023judging} to Llama3-8B~\cite{dubey2024llama}, as the latter achieves 20\% higher accuracy on MMLU. This change results in 1\% gain on spatial reasoning. We analyze that improved language reasoning may enhance spatial reasoning as well.
\item \textit{+ 3DI-Ft1M data for stage 2 instruction tuning}. We collect the new spatial-aware training set and use it in the instruction tuning stage. Impressively, this 3D-informed instruction tuning results in a 10.7\% performance improvement, highlighting the critical importance of 3D-informed data.
\item \textit{+ 3DI-Pb-1N166K data for stage 1 multimodal alignment}. We propose leveraging object-level 3D-informed annotation (e.g., 3D location and 3D orientation of objects) to align visual features with 3D awareness. 3DI-Pb-1N166K are converted from human-annotated ImageNet3D~\cite{ma2024imagenet3d}. This experiment results in an additional 3\% performance improvement, underscoring the importance of 3D-informed feature alignment.
\end{itemize}

In a nutshell, the performance gain can be summarized as 1) better architecture and 2) 3D-informed data. This roadmap leads to our \modelname.






\section{Experiments} \label{sec:exp}
In this section we start by outlining the experimental setup in Sec.~\ref{sec:exp_setup}, followed by presenting the results in Sec.~\ref{sec:exp_results}.
\subsection{Experimental setup}
\label{sec:exp_setup}
We prepare our training data as described in Fig.~\ref{fig:3di_data}, with 1M paired image and 3D-informed text for pre-pretraining and alignment, and another 1M multi-turn 3D-informed conversation for instruction tuning. Data distributions are supplemented in appendix. Our training setup is built upon LLaVA-v1.5~\cite{liu2024improved} and all hyperparameters remain unchanged unless explicitly stated otherwise.  We briefly overview the training and data setup for the ablated design choices in Sec.~\ref{sec:design_space}. All experiments are conducted in 8 Nvidia A100 GPUs. We evaluate the spatial reasoning capability of LMMs on SpatialVQA as introduced in Sec.~\ref{sec:build_spatialvqa}.

\subsection{Results} 
\label{sec:exp_results}
\textbf{Comparison with the state-of-the-art}. 
Table~\ref{tab:sota} presents the performance of our model compared to state-of-the-art proprietary systems such as GPT-4o and Claude 3.5 Sonnet, along with top open-source models.
\begin{table}[ht!]
  \centering
    \renewcommand{\arraystretch}{1.3}
    \resizebox{1.00\columnwidth}{!}{
    \Large
    \begin{tabular}{lcccc}
    \toprule
    Model & Avg. & 3D Dist. & 3D Orient. & 3D Spatial Rel. \\
    \midrule
    \textbf{\textit{Proprietary}}  & & & & \\
    GPT-4o~\cite{achiam2023gpt} & 54.0 & 59.4 & 52.4 & 50.8 \\
    GPT-4o Mini~\cite{achiam2023gpt} & 45.4 & 46.9 & 39.3& 48.5\\
    Claude 3.5 Sonnet~\cite{claude} & 49.3 & 50.8 & 44.5 & 51.5 \\
    
    \midrule
    \textbf{\textit{Open-sourced}} \\
    LLaVA-v1.5-7B~\cite{liu2024improved}  & 47.7 & 50.6 & 46.4 & 46.2 \\
    LLaVA-NeXT-8B~\cite{liu2024llavanext} & 50.6 & 63.5  & 49.2 & 41.5 \\ 
    Cambrian-8B~\cite{tong2024cambrian} & 51.7	& 59.8 & 46.4 & 48.9 \\
    \midrule
    \textbf{\textit{Spatial-centric}} \\
    SpatialVLM-13B~\cite{chen2024spatialvlm} & 52.2 &	64.3 & 48.5 & 45.8 \\
    \midrule
    \textbf{Ours} & \textbf{62.7} & \textbf{86.3} & \textbf{52.9} & \textbf{50.8} \\
    \bottomrule
  \end{tabular}
  }
  \caption{\textbf{Comparison with the state-of-the-arts} including proprietary and open source models.}
  \label{tab:sota}
\end{table}

Remarkably, our model achieves a performance of 62.7\%, outperforming the top proprietary model by 8.7\% and the best open-source model by 10.5\%. Besides, our model excels in all aspects of spatial reasoning involving 3D distance, 3D orientation and 3D spatial relationship. Moreover, our model not only enhances spatial reasoning performance, but still preserves general visual reasoning (detailed in appendix). Interestingly, although SpatialVLM~\cite{chen2024spatialvlm} (implemented in SpaceLLaVA~\cite{spacellava}) outperforms other open-source models in overall performance, it falls short in 3D orientation reasoning compared to LLaVA, due to lack of tuning with 3D orientation-specific data. We exclude SpatialRGPT from this comparison as it relies on additional inputs such as object bounding boxes, segmentation masks, and depth maps.  In contrast, our \modelname, along with LLaVA and Cambrian, focuses on learning spatial awareness directly from raw images. We will consider models with additional inputs in future work. In sum, our model achieves state-of-the-art performance on spatial reasoning, surpassing API models~\cite{achiam2023gpt, claude} and also spatial-centric VLM~\cite{chen2024spatialvlm}.

\textbf{Thorough empirical exploration of the design space}. 
We comprehensively explore the design space of our approach as shown in Table.~\ref{tab:main_results}. We analyze the results and elaborate the findings detailed below.

\textbf{Upgrading architectures enable better spatial reasoning}. In terms of architecture, integrating a mixed vision encoder can improve overall performance especially for 3D orientation. Upgrading the LLM to the more advanced LLama3-8B enhances spatial reasoning performance by 1\%, leveraging improved language reasoning capabilities.

\begin{table*}[ht!]
  \centering
  \renewcommand{\arraystretch}{1.33}
  \resizebox{1.00\textwidth}{!}{
   \Large
  \begin{tabular}{lcccc|ccccc}
    \toprule
    \multicolumn{2}{c}{Model} & Stage 0 & Stage 1 & Stage 2 & \multicolumn{4}{c}{\datasetname} \\
    Vision Encoder & LLM & Encoder pretraining & Alignment & Instruction Tuning & Avg. & Dist. & Orient. & Spatial Rel. \\
    \hline
    \textit{\textbf{Baseline} (LLaVA-v1.5)} & & & & & & & \\
    CLIP~\cite{radford2021learning} & Vicuna-7B~\cite{zheng2023judging} & -- & CC558K~\cite{liu2024improved} & Mix665K~\cite{liu2024improved} & 47.7 & 50.6 & 46.4 & 46.2 \\
    \hline
    \textbf{\textit{Mixed Encoders}} & & & & & & & \\
    \textcolor{ablatecolor}{CLIP~\cite{radford2021learning}+MAE~\cite{he2022masked}} & \textcolor{ablatecolor}{Vicuna-7B~\cite{zheng2023judging}} & \textcolor{ablatecolor}{--} & \textcolor{ablatecolor}{CC558K} & \textcolor{ablatecolor}{Mix665K} & \textcolor{ablatecolor}{47.9} & \textcolor{ablatecolor}{52.2} & \textcolor{ablatecolor}{45.7} & \textcolor{ablatecolor}{46.1} \\
    CLIP~\cite{radford2021learning}+DINOv2~\cite{oquab2023dinov2} & Vicuna-7B~\cite{zheng2023judging} & -- & CC558K & Mix665K & 48.0 & 51.9 & 46.9 & 45.6 \\
    \hline
    \textbf{\textit{Advanced LLM}} & & & & & & & \\
    CLIP~\cite{radford2021learning}+DINOv2~\cite{oquab2023dinov2} & Llama3-8B~\cite{dubey2024llama} & -- & CC558K~\cite{liu2024improved} & Mix665K~\cite{liu2024improved} & 49.0 & 56.3 & 51.7 & 41.5 \\
    \hline
    \multicolumn{2}{l}{\textbf{\textit{3D-informed Tuning for Stage 2}}} & & & & & & \\
    CLIP~\cite{radford2021learning}+DINOv2~\cite{oquab2023dinov2} & Llama3-8B~\cite{dubey2024llama} & -- & CC558K~\cite{liu2024improved} & Mix665K~\cite{liu2024improved} + 3DI-Ft1M & 59.7 & 85.2 & 52.1 & 44.8 \\
    \hline
    
    \multicolumn{3}{l}{\textbf{\textit{3D-informed Alignment for Stage 1}}} & & & & & & \\
    \textcolor{ablatecolor}{CLIP~\cite{radford2021learning}+DINOv2~\cite{oquab2023dinov2}} & \textcolor{ablatecolor}{Llama3-8B~\cite{dubey2024llama}} & \textcolor{ablatecolor}{--} & \textcolor{ablatecolor}{CC558K~\cite{liu2024improved} + 3DI-Ft1M} & \textcolor{ablatecolor}{Mix665K~\cite{liu2024improved} + 3DI-Ft1M} & \textcolor{ablatecolor}{57.5} & \textcolor{ablatecolor}{80.8} & \textcolor{ablatecolor}{46.9} & \textcolor{ablatecolor}{46.4} \\
    \textcolor{ablatecolor}{CLIP~\cite{radford2021learning}+DINOv2~\cite{oquab2023dinov2}} & \textcolor{ablatecolor}{Llama3-8B~\cite{dubey2024llama}} & \textcolor{ablatecolor}{--} & \textcolor{ablatecolor}{CC558K~\cite{liu2024improved} + 3DI-Pb-OI1M} & \textcolor{ablatecolor}{Mix665K~\cite{liu2024improved} + 3DI-Ft1M} & \textcolor{ablatecolor}{60.0} & \textcolor{ablatecolor}{82.1} & \textcolor{ablatecolor}{46.9} & \textcolor{ablatecolor}{51.6} \\
    \rowcolor{lightorange!20}
    CLIP~\cite{radford2021learning}+DINOv2~\cite{oquab2023dinov2} & Llama3-8B~\cite{dubey2024llama} & -- & CC558K~\cite{liu2024improved} + 3DI-Pb-IN166K & Mix665K~\cite{liu2024improved} + 3DI-Ft1M & \textbf{62.7} & \textbf{86.3} & \textbf{52.9} & \textbf{51.0} \\

    \hline
    \multicolumn{4}{l}{\textbf{\textcolor{ablatecolor}{\textit{3D-informed pretraining for vision encoder in stage 0}}}} & & & & & \\
    \textcolor{ablatecolor}{CLIP~\cite{radford2021learning}+DINOv2~\cite{oquab2023dinov2}} & \textcolor{ablatecolor}{Llama3-8B~\cite{dubey2024llama}} & \textcolor{ablatecolor}{3DI-Pb-OI1M (CLIP~\cite{ilharco_gabriel_2021_5143773})} & \textcolor{ablatecolor}{CC558K~\cite{liu2024improved} + 3DI-Pb-IN166K} & \textcolor{ablatecolor}{Mix665K~\cite{liu2024improved} + 3DI-Ft1M} & \textcolor{ablatecolor}{60.3} & \textcolor{ablatecolor}{85.0} 	& \textcolor{ablatecolor}{50.5} & \textcolor{ablatecolor}{47.5}  \\
    \textcolor{ablatecolor}{CLIP~\cite{radford2021learning}+DINOv2~\cite{oquab2023dinov2}} & \textcolor{ablatecolor}{Llama3-8B~\cite{dubey2024llama}} & \textcolor{ablatecolor}{3DI-Pb-OI1M (NTP~\cite{li2024llava})} & \textcolor{ablatecolor}{CC558K~\cite{liu2024improved} + 3DI-Pb-IN166K} & \textcolor{ablatecolor}{Mix665K~\cite{liu2024improved} + 3DI-Ft1M}  & \textcolor{ablatecolor}{62.6} & \textcolor{ablatecolor}{85.2} & \textcolor{ablatecolor}{52.9} & \textcolor{ablatecolor}{51.5} \\
    
    \bottomrule
  \end{tabular}
  }
  \caption{\textbf{Thorough exploration of the design space and roadmap progression.} We systematically examine the 3D-informed design space from the aspects of data, architecture and training. The non-gray rows, listed from top to bottom, illustrate the step-by-step progression of our design roadmap. Gray rows correspond to ablations across the full design space. Our \colorbox{lightorange!20}{final model} is highlighted with a light orange background. Our final model's superior performance over other variants confirms the necessity of a 3D-informed compound design.}
  \label{tab:main_results}
\end{table*}

\textbf{3D-informed alignment and tuning matter a lot.}  In terms of 3D-informed data and training, we find that 3D-informed instruction tuning with our proposed 3DI-Ft1M dataset yields a substantial performance boost of +10.7\%. To further refine the framework, we explore three data types for stage 1 multimodal alignment: curated 3DI-Ft1M, 3DI-Pb-OI1M, and 3DI-Pb-IN166K. The model trained with 3DI-Pb-IN166K performs best. We attribute this to the use of human-annotated ImageNet3D data, as opposed to the semi-automated methods employed for 3DI-Pb-OI1M. The human-annotated ImageNet3D provides more accurate 3D orientation of objects. This highlights the importance of high-quality data in the alignment stage.

\textbf{3D-informed pretraining of vision encoder?} The pre-trained CLIP vision encoder, originally designed for image-text contrastive learning, lacks 3D-awareness~\cite{el2024probing}, We hypotheszie that the vision encoder has a chance to benefit from 3D-informed data.  To verify this, we conduct 3D-informed pretraining for the vision encoder using curated 3DI-Pb-OI1M probing data. We explore both CLIP ~\cite{ilharco_gabriel_2021_5143773} and next token prediction (NTP)~\cite{liu2024visual} objectives to fine-tune Lora layers~\cite{hu2021lora} of the ViT. We observe that pre-pretraining in stage 0 reduces performance, suggesting that tuning visual foundation models may compromise feature generalizability. This aligns with prior works advocating for frozen or locked visual encoders~\cite{li2023blip, liu2023learning}.

\textbf{Roadmap progression towards the best-performing model}. 
After a thorough empirical exploration of the design space, we can devise our roadmap progressed towards the best-performing model. The non-gray rows in Table.~\ref{tab:main_results} align with the roadmap illustrated in Fig.~\ref{fig:search_roadmap}, highlighting our systematic approach to upgrade the architecture and incorporate 3D-informed data into the training process. Overall, the improved architecture achieves a 1.3\% performance gain, while 3D-informed alignment and instruction tuning further enhance performance by 13.7\%.

\begin{figure}[t]
  \centering
  \includegraphics[width=\linewidth]{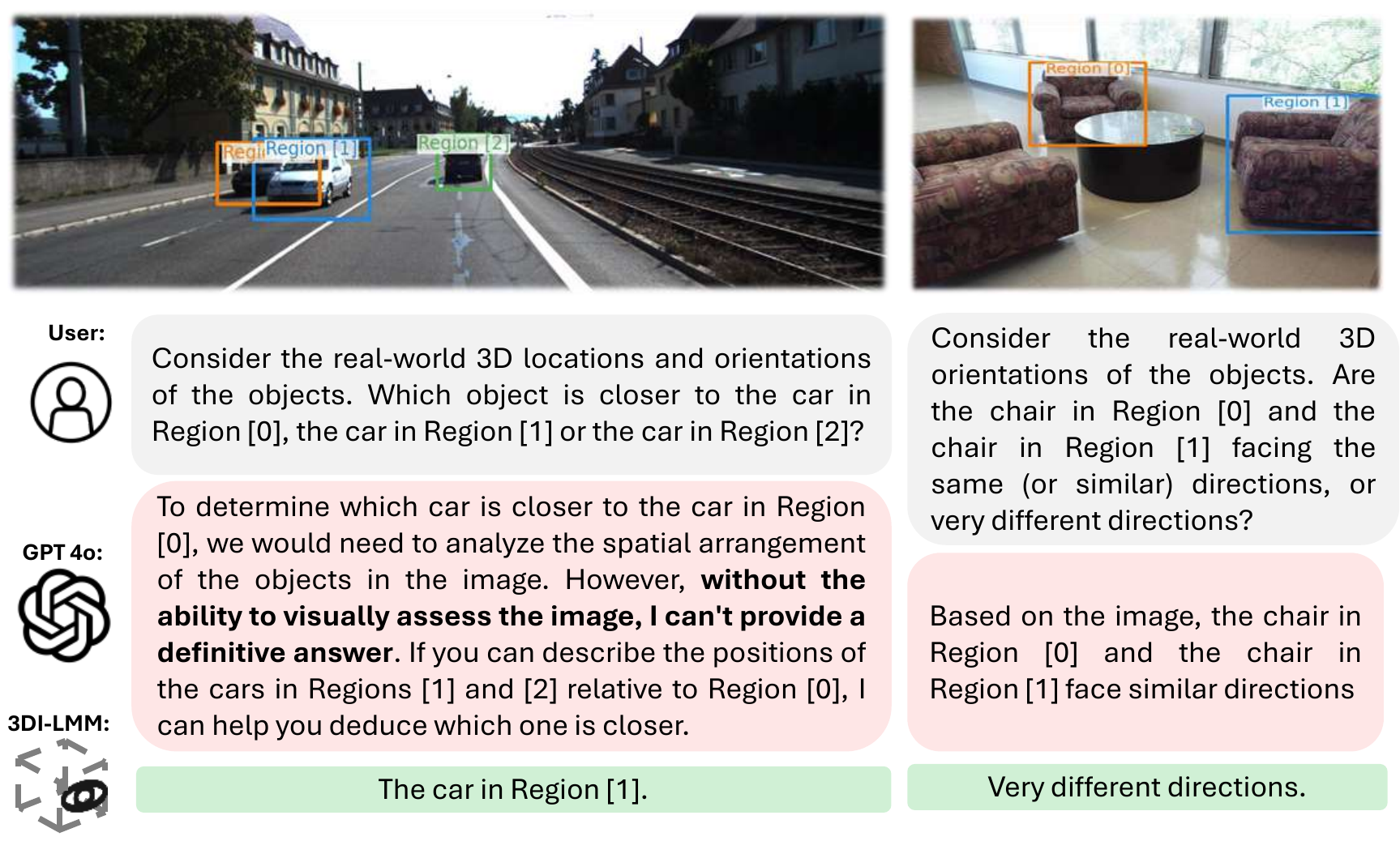}
\caption{\small Our model answers question correctly that needs accurate reasoning on \textbf{spatial distance} and \textbf{3D orientation}, while GPT-4o needs more spatial information or gives wrong answer.}
\vspace{-3mm}
   \label{fig:exp_vis}
\end{figure}


\textbf{Qualitative comparison}. 
Fig.~\ref{fig:exp_vis} illustrates a scenario involving spatial reasoning. GPT-4o struggles to reason about spatial relationships between objects due to limited ability to comprehend 3D orientations. In contrast, our model correctly predicts spatial relationships and provides accurate answers, showing superior 3D spatial reasoning capability.

\section{Conclusions}
\label{sec:conclusiosn}
In this work, we systematically enhance the 3D spatial reasoning capabilities of LMMs through a compound design approach. By introducing \modelname, we demonstrated that integrating 3D-informed data, architectural innovations, and tailored training setups significantly improves an LMM's ability to understand and reason about complex 3D spatial relationships. 
To tackle data scarcity, we develop 3D-informed training datasets focused on 3D locations, orientations and complex spatial relationship. 
Our systematic integration of these datasets with architectural and training designs provide a roadmap for optimizing LMMs for superior 3D reasoning capabilities.
Experimental results on the SpatialVQA benchmark highlight \modelname's superiority over state-of-the-art models, setting a new standard for 3D spatial reasoning. 
%
We invite the community to build upon our design roadmap and findings to push the boundaries of what LMMs can achieve in understanding the complex 3D spatial relationships of the real world.
\clearpage

\section{Acknowledgements}

WM and AY acknowledges support from ONR with N00014-23-1-2641 and ARL award W911NF2320008. JC acknowledges support from a Siebel Scholarship.
{
    \small
    \bibliographystyle{ieeenat_fullname}
    \bibliography{main}

\begin{thebibliography}{70}
\providecommand{\natexlab}[1]{#1}
\providecommand{\url}[1]{\texttt{#1}}
\expandafter\ifx\csname urlstyle\endcsname\relax
  \providecommand{\doi}[1]{doi: #1}\else
  \providecommand{\doi}{doi: \begingroup \urlstyle{rm}\Url}\fi

\bibitem[Achiam et~al.(2023)Achiam, Adler, Agarwal, Ahmad, Akkaya, Aleman, Almeida, Altenschmidt, Altman, Anadkat, et~al.]{achiam2023gpt}
Josh Achiam, Steven Adler, Sandhini Agarwal, Lama Ahmad, Ilge Akkaya, Florencia~Leoni Aleman, Diogo Almeida, Janko Altenschmidt, Sam Altman, Shyamal Anadkat, et~al.
\newblock Gpt-4 technical report.
\newblock \emph{arXiv preprint arXiv:2303.08774}, 2023.

\bibitem[AI(2024)]{spacellava}
Remyx AI.
\newblock {SpaceLLaVA}.
\newblock \url{https://huggingface.co/remyxai/SpaceLLaVA}, 2024.

\bibitem[Alayrac et~al.(2022)Alayrac, Donahue, Luc, Miech, Barr, Hasson, Lenc, Mensch, Millican, Reynolds, et~al.]{alayrac2022flamingo}
Jean-Baptiste Alayrac, Jeff Donahue, Pauline Luc, Antoine Miech, Iain Barr, Yana Hasson, Karel Lenc, Arthur Mensch, Katherine Millican, Malcolm Reynolds, et~al.
\newblock Flamingo: a visual language model for few-shot learning.
\newblock \emph{Advances in neural information processing systems}, 35:\penalty0 23716--23736, 2022.

\bibitem[Anthropic(2024)]{claude}
Anthropic.
\newblock {Claude 3.5 Sonnet}.
\newblock \url{https://www.anthropic.com/news/claude-3-5-sonnet}, 2024.

\bibitem[{Apollo Team}(2019)]{apollo}
{Apollo Team}.
\newblock Apollo syntheic dataset, 2019.

\bibitem[Azuma et~al.(2022)Azuma, Miyanishi, Kurita, and Kawanabe]{azuma2022scanqa}
Daichi Azuma, Taiki Miyanishi, Shuhei Kurita, and Motoaki Kawanabe.
\newblock Scanqa: 3d question answering for spatial scene understanding.
\newblock In \emph{proceedings of the IEEE/CVF conference on computer vision and pattern recognition}, pages 19129--19139, 2022.

\bibitem[Bai et~al.(2023{\natexlab{a}})Bai, Bai, Chu, Cui, Dang, Deng, Fan, Ge, Han, Huang, et~al.]{bai2023qwen}
Jinze Bai, Shuai Bai, Yunfei Chu, Zeyu Cui, Kai Dang, Xiaodong Deng, Yang Fan, Wenbin Ge, Yu Han, Fei Huang, et~al.
\newblock Qwen technical report.
\newblock \emph{arXiv preprint arXiv:2309.16609}, 2023{\natexlab{a}}.

\bibitem[Bai et~al.(2023{\natexlab{b}})Bai, Bai, Yang, Wang, Tan, Wang, Lin, Zhou, and Zhou]{bai2023qwenvl}
Jinze Bai, Shuai Bai, Shusheng Yang, Shijie Wang, Sinan Tan, Peng Wang, Junyang Lin, Chang Zhou, and Jingren Zhou.
\newblock Qwen-vl: A frontier large vision-language model with versatile abilities.
\newblock \emph{arXiv preprint arXiv:2308.12966}, 2023{\natexlab{b}}.

\bibitem[Baruch et~al.(2021)Baruch, Chen, Dehghan, Dimry, Feigin, Fu, Gebauer, Joffe, Kurz, Schwartz, et~al.]{baruch2021arkitscenes}
Gilad Baruch, Zhuoyuan Chen, Afshin Dehghan, Tal Dimry, Yuri Feigin, Peter Fu, Thomas Gebauer, Brandon Joffe, Daniel Kurz, Arik Schwartz, et~al.
\newblock Arkitscenes: A diverse real-world dataset for 3d indoor scene understanding using mobile rgb-d data.
\newblock \emph{arXiv preprint arXiv:2111.08897}, 2021.

\bibitem[Bhat et~al.(2023)Bhat, Birkl, Wofk, Wonka, and M{\"u}ller]{bhat2023zoedepth}
Shariq~Farooq Bhat, Reiner Birkl, Diana Wofk, Peter Wonka, and Matthias M{\"u}ller.
\newblock Zoedepth: Zero-shot transfer by combining relative and metric depth.
\newblock \emph{arXiv preprint arXiv:2302.12288}, 2023.

\bibitem[Brazil et~al.(2023)Brazil, Kumar, Straub, Ravi, Johnson, and Gkioxari]{brazil2023omni3d}
Garrick Brazil, Abhinav Kumar, Julian Straub, Nikhila Ravi, Justin Johnson, and Georgia Gkioxari.
\newblock {Omni3D}: A large benchmark and model for {3D} object detection in the wild.
\newblock In \emph{CVPR}, Vancouver, Canada, 2023. IEEE.

\bibitem[Caesar et~al.(2020)Caesar, Bankiti, Lang, Vora, Liong, Xu, Krishnan, Pan, Baldan, and Beijbom]{caesar2020nuscenes}
Holger Caesar, Varun Bankiti, Alex~H Lang, Sourabh Vora, Venice~Erin Liong, Qiang Xu, Anush Krishnan, Yu Pan, Giancarlo Baldan, and Oscar Beijbom.
\newblock nuscenes: A multimodal dataset for autonomous driving.
\newblock In \emph{Proceedings of the IEEE/CVF conference on computer vision and pattern recognition}, pages 11621--11631, 2020.

\bibitem[Cha et~al.(2024)Cha, Kang, Mun, and Roh]{cha2024honeybee}
Junbum Cha, Wooyoung Kang, Jonghwan Mun, and Byungseok Roh.
\newblock Honeybee: Locality-enhanced projector for multimodal llm.
\newblock In \emph{Proceedings of the IEEE/CVF Conference on Computer Vision and Pattern Recognition}, pages 13817--13827, 2024.

\bibitem[Chen et~al.(2024{\natexlab{a}})Chen, Xu, Kirmani, Ichter, Sadigh, Guibas, and Xia]{chen2024spatialvlm}
Boyuan Chen, Zhuo Xu, Sean Kirmani, Brain Ichter, Dorsa Sadigh, Leonidas Guibas, and Fei Xia.
\newblock Spatialvlm: Endowing vision-language models with spatial reasoning capabilities.
\newblock In \emph{Proceedings of the IEEE/CVF Conference on Computer Vision and Pattern Recognition}, pages 14455--14465, 2024{\natexlab{a}}.

\bibitem[Chen et~al.(2024{\natexlab{b}})Chen, Yu, Shen, Yuille, and Chen]{chen2024vitamin}
Jieneng Chen, Qihang Yu, Xiaohui Shen, Alan Yuille, and Liang-Chieh Chen.
\newblock Vitamin: Designing scalable vision models in the vision-language era.
\newblock In \emph{Proceedings of the IEEE/CVF Conference on Computer Vision and Pattern Recognition}, 2024{\natexlab{b}}.

\bibitem[Cheng et~al.(2024)Cheng, Yin, Fu, Guo, Yang, Kautz, Wang, and Liu]{cheng2024spatialrgpt}
An-Chieh Cheng, Hongxu Yin, Yang Fu, Qiushan Guo, Ruihan Yang, Jan Kautz, Xiaolong Wang, and Sifei Liu.
\newblock Spatialrgpt: Grounded spatial reasoning in vision language model.
\newblock \emph{arXiv preprint arXiv:2406.01584}, 2024.

\bibitem[Deng et~al.(2009)Deng, Dong, Socher, Li, Li, and Fei-Fei]{deng2009imagenet}
Jia Deng, Wei Dong, Richard Socher, Li-Jia Li, Kai Li, and Li Fei-Fei.
\newblock Imagenet: A large-scale hierarchical image database.
\newblock In \emph{2009 IEEE conference on computer vision and pattern recognition}, pages 248--255. Ieee, 2009.

\bibitem[Dubey et~al.(2024)Dubey, Jauhri, Pandey, Kadian, Al-Dahle, Letman, Mathur, Schelten, Yang, Fan, et~al.]{dubey2024llama}
Abhimanyu Dubey, Abhinav Jauhri, Abhinav Pandey, Abhishek Kadian, Ahmad Al-Dahle, Aiesha Letman, Akhil Mathur, Alan Schelten, Amy Yang, Angela Fan, et~al.
\newblock The llama 3 herd of models.
\newblock \emph{arXiv preprint arXiv:2407.21783}, 2024.

\bibitem[El~Banani et~al.(2024)El~Banani, Raj, Maninis, Kar, Li, Rubinstein, Sun, Guibas, Johnson, and Jampani]{el2024probing}
Mohamed El~Banani, Amit Raj, Kevis-Kokitsi Maninis, Abhishek Kar, Yuanzhen Li, Michael Rubinstein, Deqing Sun, Leonidas Guibas, Justin Johnson, and Varun Jampani.
\newblock Probing the 3d awareness of visual foundation models.
\newblock In \emph{Proceedings of the IEEE/CVF Conference on Computer Vision and Pattern Recognition}, pages 21795--21806, 2024.

\bibitem[Gadre et~al.(2023)Gadre, Ilharco, Fang, Hayase, Smyrnis, Nguyen, Marten, Wortsman, Ghosh, Zhang, et~al.]{gadre2023datacomp}
Samir~Yitzhak Gadre, Gabriel Ilharco, Alex Fang, Jonathan Hayase, Georgios Smyrnis, Thao Nguyen, Ryan Marten, Mitchell Wortsman, Dhruba Ghosh, Jieyu Zhang, et~al.
\newblock Datacomp: In search of the next generation of multimodal datasets.
\newblock \emph{arXiv preprint arXiv:2304.14108}, 2023.

\bibitem[Geiger et~al.(2012)Geiger, Lenz, and Urtasun]{geiger2012we}
Andreas Geiger, Philip Lenz, and Raquel Urtasun.
\newblock Are we ready for autonomous driving? the kitti vision benchmark suite.
\newblock In \emph{2012 IEEE conference on computer vision and pattern recognition}, pages 3354--3361. IEEE, 2012.

\bibitem[Goyal et~al.(2017)Goyal, Khot, Summers-Stay, Batra, and Parikh]{goyal2017making}
Yash Goyal, Tejas Khot, Douglas Summers-Stay, Dhruv Batra, and Devi Parikh.
\newblock Making the v in vqa matter: Elevating the role of image understanding in visual question answering.
\newblock In \emph{Proceedings of the IEEE conference on computer vision and pattern recognition}, pages 6904--6913, 2017.

\bibitem[He et~al.(2022)He, Chen, Xie, Li, Doll{\'a}r, and Girshick]{he2022masked}
Kaiming He, Xinlei Chen, Saining Xie, Yanghao Li, Piotr Doll{\'a}r, and Ross Girshick.
\newblock Masked autoencoders are scalable vision learners.
\newblock In \emph{Proceedings of the IEEE/CVF conference on computer vision and pattern recognition}, pages 16000--16009, 2022.

\bibitem[Hu et~al.(2021)Hu, Shen, Wallis, Allen-Zhu, Li, Wang, Wang, and Chen]{hu2021lora}
Edward~J Hu, Yelong Shen, Phillip Wallis, Zeyuan Allen-Zhu, Yuanzhi Li, Shean Wang, Lu Wang, and Weizhu Chen.
\newblock Lora: Low-rank adaptation of large language models.
\newblock \emph{arXiv preprint arXiv:2106.09685}, 2021.

\bibitem[Ilharco et~al.(2021)Ilharco, Wortsman, Wightman, Gordon, Carlini, Taori, Dave, Shankar, Namkoong, Miller, Hajishirzi, Farhadi, and Schmidt]{ilharco_gabriel_2021_5143773}
Gabriel Ilharco, Mitchell Wortsman, Ross Wightman, Cade Gordon, Nicholas Carlini, Rohan Taori, Achal Dave, Vaishaal Shankar, Hongseok Namkoong, John Miller, Hannaneh Hajishirzi, Ali Farhadi, and Ludwig Schmidt.
\newblock Openclip, 2021.
\newblock If you use this software, please cite it as below.

\bibitem[Jesslen et~al.(2024)Jesslen, Zhang, Wang, Ma, Yuille, and Kortylewski]{jesslen2024novum}
Artur Jesslen, Guofeng Zhang, Angtian Wang, Wufei Ma, Alan Yuille, and Adam Kortylewski.
\newblock Novum: Neural object volumes for robust object classification.
\newblock In \emph{European Conference on Computer Vision}, pages 264--281. Springer, 2024.

\bibitem[Jia et~al.(2021)Jia, Yang, Xia, Chen, Parekh, Pham, Le, Sung, Li, and Duerig]{jia2021scaling}
Chao Jia, Yinfei Yang, Ye Xia, Yi-Ting Chen, Zarana Parekh, Hieu Pham, Quoc Le, Yun-Hsuan Sung, Zhen Li, and Tom Duerig.
\newblock Scaling up visual and vision-language representation learning with noisy text supervision.
\newblock In \emph{ICML}, 2021.

\bibitem[Jiang et~al.(2023)Jiang, Sablayrolles, Mensch, Bamford, Chaplot, Casas, Bressand, Lengyel, Lample, Saulnier, et~al.]{jiang2023mistral}
Albert~Q Jiang, Alexandre Sablayrolles, Arthur Mensch, Chris Bamford, Devendra~Singh Chaplot, Diego de~las Casas, Florian Bressand, Gianna Lengyel, Guillaume Lample, Lucile Saulnier, et~al.
\newblock Mistral 7b.
\newblock \emph{arXiv preprint arXiv:2310.06825}, 2023.

\bibitem[Jin et~al.(2023)Jin, Zhang, Hold-Geoffroy, Wang, Blackburn-Matzen, Sticha, and Fouhey]{jin2023perspective}
Linyi Jin, Jianming Zhang, Yannick Hold-Geoffroy, Oliver Wang, Kevin Blackburn-Matzen, Matthew Sticha, and David~F Fouhey.
\newblock Perspective fields for single image camera calibration.
\newblock In \emph{Proceedings of the IEEE/CVF Conference on Computer Vision and Pattern Recognition}, pages 17307--17316, 2023.

\bibitem[Kirillov et~al.(2023{\natexlab{a}})Kirillov, Mintun, Ravi, Mao, Rolland, Gustafson, Xiao, Whitehead, Berg, Lo, Doll{\'a}r, and Girshick]{kirillov2023segany}
Alexander Kirillov, Eric Mintun, Nikhila Ravi, Hanzi Mao, Chloe Rolland, Laura Gustafson, Tete Xiao, Spencer Whitehead, Alexander~C. Berg, Wan-Yen Lo, Piotr Doll{\'a}r, and Ross Girshick.
\newblock Segment anything.
\newblock \emph{arXiv:2304.02643}, 2023{\natexlab{a}}.

\bibitem[Kirillov et~al.(2023{\natexlab{b}})Kirillov, Mintun, Ravi, Mao, Rolland, Gustafson, Xiao, Whitehead, Berg, Lo, et~al.]{kirillov2023segment}
Alexander Kirillov, Eric Mintun, Nikhila Ravi, Hanzi Mao, Chloe Rolland, Laura Gustafson, Tete Xiao, Spencer Whitehead, Alexander~C Berg, Wan-Yen Lo, et~al.
\newblock Segment anything.
\newblock In \emph{Proceedings of the IEEE/CVF International Conference on Computer Vision}, pages 4015--4026, 2023{\natexlab{b}}.

\bibitem[Krishna et~al.(2017)Krishna, Zhu, Groth, Johnson, Hata, Kravitz, Chen, Kalantidis, Li, Shamma, et~al.]{krishna2017visual}
Ranjay Krishna, Yuke Zhu, Oliver Groth, Justin Johnson, Kenji Hata, Joshua Kravitz, Stephanie Chen, Yannis Kalantidis, Li-Jia Li, David~A Shamma, et~al.
\newblock Visual genome: Connecting language and vision using crowdsourced dense image annotations.
\newblock \emph{International journal of computer vision}, 123:\penalty0 32--73, 2017.

\bibitem[Kuznetsova et~al.(2020)Kuznetsova, Rom, Alldrin, Uijlings, Krasin, Pont-Tuset, Kamali, Popov, Malloci, Kolesnikov, et~al.]{kuznetsova2020open}
Alina Kuznetsova, Hassan Rom, Neil Alldrin, Jasper Uijlings, Ivan Krasin, Jordi Pont-Tuset, Shahab Kamali, Stefan Popov, Matteo Malloci, Alexander Kolesnikov, et~al.
\newblock The open images dataset v4: Unified image classification, object detection, and visual relationship detection at scale.
\newblock \emph{International journal of computer vision}, 128\penalty0 (7):\penalty0 1956--1981, 2020.

\bibitem[Li et~al.(2024{\natexlab{a}})Li, Zhang, Guo, Zhang, Li, Zhang, Zhang, Li, Liu, and Li]{li2024llava}
Bo Li, Yuanhan Zhang, Dong Guo, Renrui Zhang, Feng Li, Hao Zhang, Kaichen Zhang, Yanwei Li, Ziwei Liu, and Chunyuan Li.
\newblock Llava-onevision: Easy visual task transfer.
\newblock \emph{arXiv preprint arXiv:2408.03326}, 2024{\natexlab{a}}.

\bibitem[Li et~al.(2022)Li, Li, Xiong, and Hoi]{li2022blip}
Junnan Li, Dongxu Li, Caiming Xiong, and Steven Hoi.
\newblock Blip: Bootstrapping language-image pre-training for unified vision-language understanding and generation.
\newblock In \emph{ICML}, 2022.

\bibitem[Li et~al.(2023)Li, Li, Savarese, and Hoi]{li2023blip}
Junnan Li, Dongxu Li, Silvio Savarese, and Steven Hoi.
\newblock Blip-2: Bootstrapping language-image pre-training with frozen image encoders and large language models.
\newblock In \emph{International conference on machine learning}, pages 19730--19742. PMLR, 2023.

\bibitem[Li et~al.(2024{\natexlab{b}})Li, Tu, Hui, Wang, Zhao, Xiao, Ren, Mei, Liu, Zheng, et~al.]{li2024if}
Xianhang Li, Haoqin Tu, Mude Hui, Zeyu Wang, Bingchen Zhao, Junfei Xiao, Sucheng Ren, Jieru Mei, Qing Liu, Huangjie Zheng, et~al.
\newblock What if we recaption billions of web images with llama-3?
\newblock \emph{arXiv preprint arXiv:2406.08478}, 2024{\natexlab{b}}.

\bibitem[Liu et~al.(2023{\natexlab{a}})Liu, Son, Yang, Liu, Gao, Lee, and Li]{liu2023learning}
Haotian Liu, Kilho Son, Jianwei Yang, Ce Liu, Jianfeng Gao, Yong~Jae Lee, and Chunyuan Li.
\newblock Learning customized visual models with retrieval-augmented knowledge.
\newblock In \emph{Proceedings of the IEEE/CVF Conference on Computer Vision and Pattern Recognition}, pages 15148--15158, 2023{\natexlab{a}}.

\bibitem[Liu et~al.(2024{\natexlab{a}})Liu, Li, Li, and Lee]{liu2024improved}
Haotian Liu, Chunyuan Li, Yuheng Li, and Yong~Jae Lee.
\newblock Improved baselines with visual instruction tuning.
\newblock In \emph{Proceedings of the IEEE/CVF Conference on Computer Vision and Pattern Recognition}, pages 26296--26306, 2024{\natexlab{a}}.

\bibitem[Liu et~al.(2024{\natexlab{b}})Liu, Li, Li, Li, Zhang, Shen, and Lee]{liu2024llavanext}
Haotian Liu, Chunyuan Li, Yuheng Li, Bo Li, Yuanhan Zhang, Sheng Shen, and Yong~Jae Lee.
\newblock Llava-next: Improved reasoning, ocr, and world knowledge, 2024{\natexlab{b}}.

\bibitem[Liu et~al.(2024{\natexlab{c}})Liu, Li, Wu, and Lee]{liu2024visual}
Haotian Liu, Chunyuan Li, Qingyang Wu, and Yong~Jae Lee.
\newblock Visual instruction tuning.
\newblock \emph{Advances in neural information processing systems}, 36, 2024{\natexlab{c}}.

\bibitem[Liu et~al.(2023{\natexlab{b}})Liu, Zeng, Ren, Li, Zhang, Yang, Jiang, Li, Yang, Su, et~al.]{liu2023grounding}
Shilong Liu, Zhaoyang Zeng, Tianhe Ren, Feng Li, Hao Zhang, Jie Yang, Qing Jiang, Chunyuan Li, Jianwei Yang, Hang Su, et~al.
\newblock Grounding dino: Marrying dino with grounded pre-training for open-set object detection.
\newblock \emph{arXiv preprint arXiv:2303.05499}, 2023{\natexlab{b}}.

\bibitem[Lu et~al.(2024)Lu, Liu, Zhang, Wang, Dong, Liu, Sun, Ren, Li, Yang, et~al.]{lu2024deepseek}
Haoyu Lu, Wen Liu, Bo Zhang, Bingxuan Wang, Kai Dong, Bo Liu, Jingxiang Sun, Tongzheng Ren, Zhuoshu Li, Hao Yang, et~al.
\newblock Deepseek-vl: towards real-world vision-language understanding.
\newblock \emph{arXiv preprint arXiv:2403.05525}, 2024.

\bibitem[Ma et~al.(2022)Ma, Wang, Yuille, and Kortylewski]{ma2022robust}
Wufei Ma, Angtian Wang, Alan Yuille, and Adam Kortylewski.
\newblock Robust category-level 6d pose estimation with coarse-to-fine rendering of neural features.
\newblock In \emph{European Conference on Computer Vision}, pages 492--508. Springer, 2022.

\bibitem[Ma et~al.(2024)Ma, Zeng, Zhang, Liu, Zhang, Kortylewski, Liu, and Yuille]{ma2024imagenet3d}
Wufei Ma, Guanning Zeng, Guofeng Zhang, Qihao Liu, Letian Zhang, Adam Kortylewski, Yaoyao Liu, and Alan Yuille.
\newblock Imagenet3d: Towards general-purpose object-level 3d understanding.
\newblock In \emph{NeurIPS}, 2024.

\bibitem[McKinzie et~al.(2024)McKinzie, Gan, Fauconnier, Dodge, Zhang, Dufter, Shah, Du, Peng, Weers, et~al.]{mckinzie2024mm1}
Brandon McKinzie, Zhe Gan, Jean-Philippe Fauconnier, Sam Dodge, Bowen Zhang, Philipp Dufter, Dhruti Shah, Xianzhi Du, Futang Peng, Floris Weers, et~al.
\newblock Mm1: Methods, analysis \& insights from multimodal llm pre-training.
\newblock \emph{arXiv preprint arXiv:2403.09611}, 2024.

\bibitem[Oquab et~al.(2024)Oquab, Darcet, Moutakanni, Vo, Szafraniec, Khalidov, Fernandez, Haziza, Massa, El-Nouby, et~al.]{oquab2023dinov2}
Maxime Oquab, Timoth{\'e}e Darcet, Th{\'e}o Moutakanni, Huy Vo, Marc Szafraniec, Vasil Khalidov, Pierre Fernandez, Daniel Haziza, Francisco Massa, Alaaeldin El-Nouby, et~al.
\newblock Dinov2: Learning robust visual features without supervision.
\newblock \emph{TMLR}, 2024.

\bibitem[Radford et~al.(2021)Radford, Kim, Hallacy, Ramesh, Goh, Agarwal, Sastry, Askell, Mishkin, Clark, et~al.]{radford2021learning}
Alec Radford, Jong~Wook Kim, Chris Hallacy, Aditya Ramesh, Gabriel Goh, Sandhini Agarwal, Girish Sastry, Amanda Askell, Pamela Mishkin, Jack Clark, et~al.
\newblock Learning transferable visual models from natural language supervision.
\newblock In \emph{ICML}, 2021.

\bibitem[Rajabi and Kosecka(2024)]{rajabi2024gsr}
Navid Rajabi and Jana Kosecka.
\newblock Gsr-bench: A benchmark for grounded spatial reasoning evaluation via multimodal llms.
\newblock \emph{arXiv preprint arXiv:2406.13246}, 2024.

\bibitem[Roberts et~al.(2021)Roberts, Ramapuram, Ranjan, Kumar, Bautista, Paczan, Webb, and Susskind]{hypersim}
Mike Roberts, Jason Ramapuram, Anurag Ranjan, Atulit Kumar, Miguel~Angel Bautista, Nathan Paczan, Russ Webb, and Joshua~M. Susskind.
\newblock {Hypersim}: {A} photorealistic synthetic dataset for holistic indoor scene understanding.
\newblock In \emph{International Conference on Computer Vision (ICCV) 2021}, 2021.

\bibitem[Rombach et~al.(2022)Rombach, Blattmann, Lorenz, Esser, and Ommer]{rombach2022high}
Robin Rombach, Andreas Blattmann, Dominik Lorenz, Patrick Esser, and Bj{\"o}rn Ommer.
\newblock High-resolution image synthesis with latent diffusion models.
\newblock In \emph{Proceedings of the IEEE/CVF conference on computer vision and pattern recognition}, pages 10684--10695, 2022.

\bibitem[Schuhmann et~al.(2022)Schuhmann, Beaumont, Vencu, Gordon, Wightman, Cherti, Coombes, Katta, Mullis, Wortsman, et~al.]{schuhmann2022laion}
Christoph Schuhmann, Romain Beaumont, Richard Vencu, Cade Gordon, Ross Wightman, Mehdi Cherti, Theo Coombes, Aarush Katta, Clayton Mullis, Mitchell Wortsman, et~al.
\newblock Laion-5b: An open large-scale dataset for training next generation image-text models.
\newblock \emph{Advances in Neural Information Processing Systems}, 35:\penalty0 25278--25294, 2022.

\bibitem[Sharma et~al.(2018)Sharma, Ding, Goodman, and Soricut]{sharma2018conceptual}
Piyush Sharma, Nan Ding, Sebastian Goodman, and Radu Soricut.
\newblock Conceptual captions: A cleaned, hypernymed, image alt-text dataset for automatic image captioning.
\newblock In \emph{ACL}, 2018.

\bibitem[Song et~al.(2015)Song, Lichtenberg, and Xiao]{song2015sun}
Shuran Song, Samuel~P Lichtenberg, and Jianxiong Xiao.
\newblock Sun rgb-d: A rgb-d scene understanding benchmark suite.
\newblock In \emph{Proceedings of the IEEE conference on computer vision and pattern recognition}, pages 567--576, 2015.

\bibitem[Spelke and Kinzler(2007)]{spelke2007core}
Elizabeth~S Spelke and Katherine~D Kinzler.
\newblock Core knowledge.
\newblock \emph{Developmental science}, 10\penalty0 (1):\penalty0 89--96, 2007.

\bibitem[Sun et~al.(2017)Sun, Shrivastava, Singh, and Gupta]{sun2017revisiting}
Chen Sun, Abhinav Shrivastava, Saurabh Singh, and Abhinav Gupta.
\newblock Revisiting unreasonable effectiveness of data in deep learning era.
\newblock In \emph{Proceedings of the IEEE international conference on computer vision}, pages 843--852, 2017.

\bibitem[Team et~al.(2023)Team, Anil, Borgeaud, Wu, Alayrac, Yu, Soricut, Schalkwyk, Dai, Hauth, et~al.]{team2023gemini}
Gemini Team, Rohan Anil, Sebastian Borgeaud, Yonghui Wu, Jean-Baptiste Alayrac, Jiahui Yu, Radu Soricut, Johan Schalkwyk, Andrew~M Dai, Anja Hauth, et~al.
\newblock Gemini: a family of highly capable multimodal models.
\newblock \emph{arXiv preprint arXiv:2312.11805}, 2023.

\bibitem[Tong et~al.(2024)Tong, Brown, Wu, Woo, Middepogu, Akula, Yang, Yang, Iyer, Pan, et~al.]{tong2024cambrian}
Shengbang Tong, Ellis Brown, Penghao Wu, Sanghyun Woo, Manoj Middepogu, Sai~Charitha Akula, Jihan Yang, Shusheng Yang, Adithya Iyer, Xichen Pan, et~al.
\newblock Cambrian-1: A fully open, vision-centric exploration of multimodal llms.
\newblock \emph{arXiv preprint arXiv:2406.16860}, 2024.

\bibitem[Touvron et~al.(2023)Touvron, Martin, Stone, Albert, Almahairi, Babaei, Bashlykov, Batra, Bhargava, Bhosale, Bikel, Blecher, Ferrer, Chen, Cucurull, Esiobu, Fernandes, Fu, Fu, Fuller, Gao, Goswami, Goyal, Hartshorn, Hosseini, Hou, Inan, Kardas, Kerkez, Khabsa, Kloumann, Korenev, Koura, Lachaux, Lavril, Lee, Liskovich, Lu, Mao, Martinet, Mihaylov, Mishra, Molybog, Nie, Poulton, Reizenstein, Rungta, Saladi, Schelten, Silva, Smith, Subramanian, Tan, Tang, Taylor, Williams, Kuan, Xu, Yan, Zarov, Zhang, Fan, Kambadur, Narang, Rodriguez, Stojnic, Edunov, and Scialom]{touvron2023llama2}
Hugo Touvron, Louis Martin, Kevin Stone, Peter Albert, Amjad Almahairi, Yasmine Babaei, Nikolay Bashlykov, Soumya Batra, Prajjwal Bhargava, Shruti Bhosale, Dan Bikel, Lukas Blecher, Cristian~Canton Ferrer, Moya Chen, Guillem Cucurull, David Esiobu, Jude Fernandes, Jeremy Fu, Wenyin Fu, Brian Fuller, Cynthia Gao, Vedanuj Goswami, Naman Goyal, Anthony Hartshorn, Saghar Hosseini, Rui Hou, Hakan Inan, Marcin Kardas, Viktor Kerkez, Madian Khabsa, Isabel Kloumann, Artem Korenev, Punit~Singh Koura, Marie-Anne Lachaux, Thibaut Lavril, Jenya Lee, Diana Liskovich, Yinghai Lu, Yuning Mao, Xavier Martinet, Todor Mihaylov, Pushkar Mishra, Igor Molybog, Yixin Nie, Andrew Poulton, Jeremy Reizenstein, Rashi Rungta, Kalyan Saladi, Alan Schelten, Ruan Silva, Eric~Michael Smith, Ranjan Subramanian, Xiaoqing~Ellen Tan, Binh Tang, Ross Taylor, Adina Williams, Jian~Xiang Kuan, Puxin Xu, Zheng Yan, Iliyan Zarov, Yuchen Zhang, Angela Fan, Melanie Kambadur, Sharan Narang, Aurelien Rodriguez, Robert Stojnic, Sergey Edunov, and Thomas
  Scialom.
\newblock Llama 2: Open foundation and fine-tuned chat models.
\newblock \emph{arXiv preprint arXiv:2307.09288}, 2023.

\bibitem[Wang et~al.(2024{\natexlab{a}})Wang, Ma, Li, Kortylewski, and Yuille]{wang20243d}
Xingrui Wang, Wufei Ma, Zhuowan Li, Adam Kortylewski, and Alan~L Yuille.
\newblock 3d-aware visual question answering about parts, poses and occlusions.
\newblock \emph{Advances in Neural Information Processing Systems}, 36, 2024{\natexlab{a}}.

\bibitem[Wang et~al.(2024{\natexlab{b}})Wang, Ma, Wang, Chen, Kortylewski, and Yuille]{wang2024compositional}
Xingrui Wang, Wufei Ma, Angtian Wang, Shuo Chen, Adam Kortylewski, and Alan Yuille.
\newblock Compositional 4d dynamic scenes understanding with physics priors for video question answering.
\newblock \emph{arXiv preprint arXiv:2406.00622}, 2024{\natexlab{b}}.

\bibitem[Yang et~al.(2024{\natexlab{a}})Yang, Kang, Huang, Xu, Feng, and Zhao]{depthanything}
Lihe Yang, Bingyi Kang, Zilong Huang, Xiaogang Xu, Jiashi Feng, and Hengshuang Zhao.
\newblock Depth anything: Unleashing the power of large-scale unlabeled data.
\newblock In \emph{CVPR}, 2024{\natexlab{a}}.

\bibitem[Yang et~al.(2024{\natexlab{b}})Yang, Kang, Huang, Xu, Feng, and Zhao]{yang2024depth}
Lihe Yang, Bingyi Kang, Zilong Huang, Xiaogang Xu, Jiashi Feng, and Hengshuang Zhao.
\newblock Depth anything: Unleashing the power of large-scale unlabeled data.
\newblock In \emph{Proceedings of the IEEE/CVF Conference on Computer Vision and Pattern Recognition}, pages 10371--10381, 2024{\natexlab{b}}.

\bibitem[Ye et~al.(2021)Ye, Chen, Han, and Liao]{ye20213d}
Shuquan Ye, Dongdong Chen, Songfang Han, and Jing Liao.
\newblock 3d question answering.
\newblock \emph{arXiv preprint arXiv:2112.08359}, 2021.

\bibitem[Yu et~al.(2022)Yu, Wang, Vasudevan, Yeung, Seyedhosseini, and Wu]{yu2022coca}
Jiahui Yu, Zirui Wang, Vijay Vasudevan, Legg Yeung, Mojtaba Seyedhosseini, and Yonghui Wu.
\newblock Coca: Contrastive captioners are image-text foundation models.
\newblock \emph{arXiv preprint arXiv:2205.01917}, 2022.

\bibitem[Zhang et~al.(2023)Zhang, Huang, Ma, Li, Luo, Xie, Qin, Luo, Li, Liu, et~al.]{zhang2023recognize}
Youcai Zhang, Xinyu Huang, Jinyu Ma, Zhaoyang Li, Zhaochuan Luo, Yanchun Xie, Yuzhuo Qin, Tong Luo, Yaqian Li, Shilong Liu, et~al.
\newblock Recognize anything: A strong image tagging model.
\newblock \emph{arXiv preprint arXiv:2306.03514}, 2023.

\bibitem[Zhang et~al.(2024)Zhang, Huang, Ma, Li, Luo, Xie, Qin, Luo, Li, Liu, et~al.]{zhang2024recognize}
Youcai Zhang, Xinyu Huang, Jinyu Ma, Zhaoyang Li, Zhaochuan Luo, Yanchun Xie, Yuzhuo Qin, Tong Luo, Yaqian Li, Shilong Liu, et~al.
\newblock Recognize anything: A strong image tagging model.
\newblock In \emph{Proceedings of the IEEE/CVF Conference on Computer Vision and Pattern Recognition}, pages 1724--1732, 2024.

\bibitem[Zheng et~al.(2023)Zheng, Chiang, Sheng, Zhuang, Wu, Zhuang, Lin, Li, Li, Xing, et~al.]{zheng2023judging}
Lianmin Zheng, Wei-Lin Chiang, Ying Sheng, Siyuan Zhuang, Zhanghao Wu, Yonghao Zhuang, Zi Lin, Zhuohan Li, Dacheng Li, Eric Xing, et~al.
\newblock Judging llm-as-a-judge with mt-bench and chatbot arena.
\newblock \emph{Advances in Neural Information Processing Systems}, 36:\penalty0 46595--46623, 2023.

\bibitem[Zhou et~al.(2021)Zhou, Wei, Wang, Shen, Xie, Yuille, and Kong]{zhou2021ibot}
Jinghao Zhou, Chen Wei, Huiyu Wang, Wei Shen, Cihang Xie, Alan Yuille, and Tao Kong.
\newblock ibot: Image bert pre-training with online tokenizer.
\newblock \emph{arXiv preprint arXiv:2111.07832}, 2021.

\bibitem[Zhu et~al.(2023)Zhu, Kumar, Hu, and Liu]{zhu2023tame}
Shengjie Zhu, Abhinav Kumar, Masa Hu, and Xiaoming Liu.
\newblock Tame a wild camera: In-the-wild monocular camera calibration.
\newblock In \emph{NeurIPS}, 2023.

\end{thebibliography}
}

\clearpage
\setcounter{page}{1}
\setcounter{section}{0}
\renewcommand{\thesection}{\Alph{section}}
\maketitlesupplementary

\section{3D-Informed Data generation} \label{sec:supp-datagen}

Previous studies~\cite{chen2024spatialvlm,cheng2024spatialrgpt} exploited visual foundation models for depth estimation, \eg, DepthAnythingv2~\cite{depthanything}, camera calibration \eg, WildCamera~\cite{zhu2023tame} and PerspectiveFields~\cite{jin2023perspective}, and object tagging and grounding, \eg, RAM~\cite{zhang2023recognize} and SAM~\cite{kirillov2023segany}. By back projecting the points into 3D camera space, they estimated an axis-aligned 3D bounding box with category name or caption for each object. This enabled SpatialVLM~\cite{chen2024spatialvlm} and SpatialRGPT~\cite{cheng2024spatialrgpt} to explore 3D distance-based relationships or 2D spatial reasoning questions, \eg, which is further left in the 2D image plane. Meanwhile, more complex 3D spatial reasoning questions are left unexplored, due to the lack of 3D orientation knowledge.

We address this key limitation by exploiting existing datasets with 3D pose annotations~\cite{ma2024imagenet3d} or synthetic data with groundtruth 3D bounding boxes~\cite{apollo}. Besides 3D poses computed from the 3D annotations in \cite{apollo}, we adopt a 3D pose estimator pretrained on \cite{ma2024imagenet3d} and predict 3D orientations for rigid objects grounded in unannotated natural images. We aggregate all collected 3D information and produce oriented 3D bounding boxes, rather than axis-aligned bounding boxes considered in previous works~\cite{chen2024spatialvlm,cheng2024spatialrgpt}. Lastly we generate diverse and rich 3D-informed data about various 3D spatial relationships.

\paragraph{3D-informed 3D probing data.} Motivated by previous linear probing experiments on 3D awareness of visual foundation models~\cite{el2024probing,ma2024imagenet3d}, we propose to exploit 3D-informed probing data and extract visual features with rich 3D awareness at the feature alignment stage. Specifically, 3D-informed probing data consists of fundamental 3D questions, \eg, depth of an object, distances between two objects, and azimuth/elevation rotations. We pretrain the visual connector only while freezing other modules, such that the visual tokens would contain rich 3D information that benefit subsequent 3D spatial reasoning. Depending on the dataset used, we present: (i) \textit{3DI-Pb-IN166K} with existing 3D annotations in ImageNet3D~\cite{ma2024imagenet3d}; and (ii) \textit{3DI-Pb-OI1M} with images from the OpenImages dataset~\cite{kuznetsova2020open} and 3D orientations estimated with a 3D pose estimator.

\paragraph{3D-informed instruction tuning data.} We further present 3D-informed instruction tuning data, with 1 million question-answer data about 3D spatial relationships on images from \cite{kuznetsova2020open}, \ie, \textit{3DI-Ft1M}. By finetuning on our 3D-informed instruction tuning data, LMMs learn to aggregate the 3D-aware information from the visual tokens and to perform 3D spatial reasoning. This highlights the necessity of adopting 3D-informed data at both the pretraining and finetuning stage, which in our experiments, we find that LMMs pretrained with our 3D-informed probing data and finetuned with our 3D-informed instruction tuning data achieves the best 3D spatial reasoning capabilities.

\begin{figure*}[t]
  \centering
  \includegraphics[width=\textwidth]{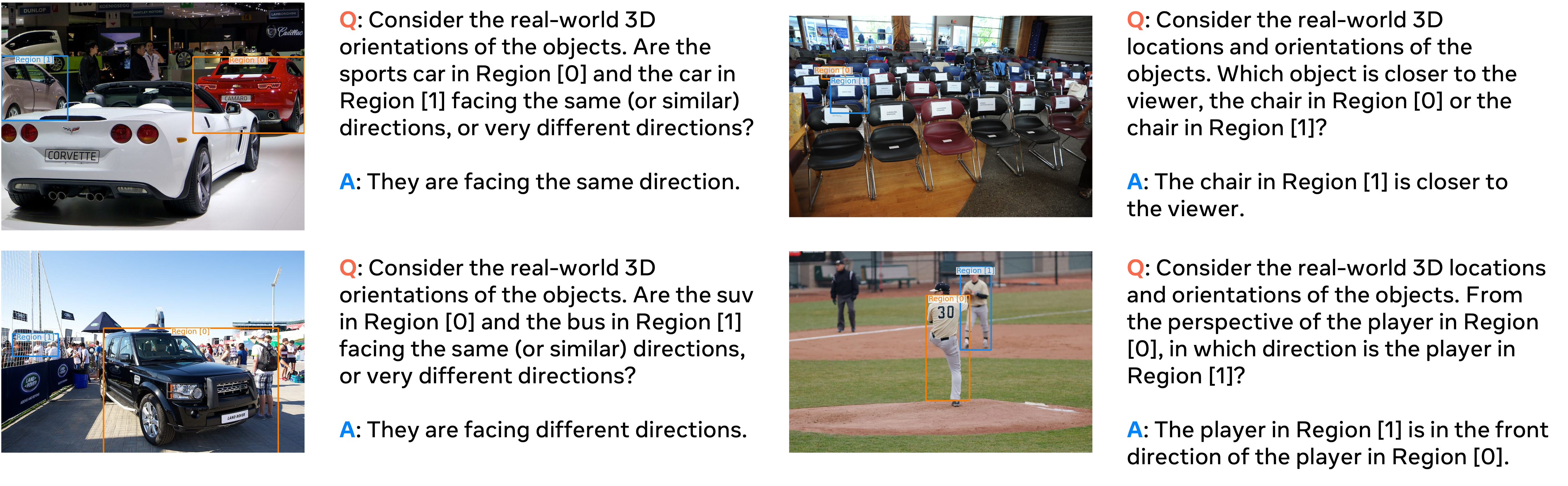}
\caption{\small Qualitative examples of our 3DI-Ft1M data.}
   \label{fig:supp_qualitative}
\end{figure*}

\paragraph{Qualitative examples.} We present some qualitative examples of our 3DI-Ft1M data in \cref{fig:supp_qualitative}.

\section{\datasetname} \label{sec:spatialvqa}

Previous spatial reasoning benchmarks were either built on 3D scans rather than images~\cite{azuma2022scanqa,ye20213d}, or focused on 2D spatial relationships~\cite{cheng2024spatialrgpt,chen2024spatialvlm}, \eg, left or right in the image plane, or only on distance spatial relationships~\cite{tong2024cambrian}.

To quantitatively study the 3D spatial reasoning capabilities of LMMs, specifically on questions about object 3D orientations or questions that require reasoning over 3D locations and 3D orientations, we build a new evaluation dataset, \ie, \datasetname. Please refer to \cref{fig:spatialvqa} for some qualitative examples of \datasetname.

\paragraph{Dataset generation.} We follow previous works~\cite{cheng2024spatialrgpt,tong2024cambrian} and develop \datasetname{} based on the 3D annotations in Omni3D~\cite{brazil2023omni3d}, with images from both urban~\cite{caesar2020nuscenes,geiger2012we} and indoor scenes~\cite{song2015sun,baruch2021arkitscenes,hypersim}. Specifically, we compute 3D locations, 3D orientations, 3D distances, and spatial relationships from the object-level 3D bounding box annotations. We generate visual question-answer pairs based on the 3D annotations using pre-defined rules for each question types.

\paragraph{Question types.} Our \datasetname{} consists of three question types: (i) distance questions that can be answered from a 3D-awareness of distance only, (ii) orientation questions that require estimating objects' 3D orientations, and (iii) spatial relationship questions that require 3D spatial reasoning over various 3D information.
\begin{enumerate}
    \item \textbf{Closer to camera [distance]:} determining which of the two objects is closer to the camera (viewer).
    \item \textbf{Closer to object [distance]:} determining which of the two objects is closer to a third object.
    \item \textbf{Facing camera [orientation]:} determining which side of the object is facing towards the camera (viewer).
    \item \textbf{Facing object [orientation]:} determining which of the two object is the third object facing towards.
    \item \textbf{Same direction [spatial relationship]:} determining if two objects are facing the same or different directions.
    \item \textbf{Higher [spatial relationship]:} determining which of the two objects has a higher 3D location.
    \item \textbf{On which side [spatial relationship]:} determining an object is on which side of another object, \eg, front, left, \etc
\end{enumerate}

\paragraph{Dataset statistics.} Our \datasetname{} include a total of 1,323 questions, with around 240 questions in each question type. Moreover, the answers to the questions are roughly balanced, \eg, for the 240 questions asking if two objects are facing the same directions, 120 questions have ``yes'' as the answer and 120 questions have ``no'' as the answer. Certain answers have a lower frequency, due to their natural scarcity in real images, \eg, the bottom of an object is facing towards the camera.

\section{Public Release}



See ``Data \& Code'' section on our \href{https://3d-spatial-reasoning.github.io/spatial-llm/}{project page}.

\end{document}